\def\year{2021}\relax
\documentclass[letterpaper]{article} % DO NOT CHANGE THIS
\usepackage{aaai21}  % DO NOT CHANGE THIS
\usepackage{times}  % DO NOT CHANGE THIS
\usepackage{helvet} % DO NOT CHANGE THIS
\usepackage{courier}  % DO NOT CHANGE THIS
\usepackage[hyphens]{url}  % DO NOT CHANGE THIS
\usepackage{graphicx} % DO NOT CHANGE THIS
\usepackage[switch]{lineno}  
\def\UrlFont{\rm}  % DO NOT CHANGE THIS
\usepackage{natbib}  % DO NOT CHANGE THIS AND DO NOT ADD ANY OPTIONS TO IT
\usepackage{caption} % DO NOT CHANGE THIS AND DO NOT ADD ANY OPTIONS TO IT
\usepackage{latexsym}
\usepackage{fancyhdr,graphicx,amsmath,amssymb}
\usepackage{multirow}
\usepackage{tabularx}
\usepackage[ruled,vlined]{algorithm2e}
\usepackage[colorinlistoftodos]{todonotes}
\usepackage{color,soul}
\usepackage{fancybox}
\usepackage{booktabs}
\usepackage{subcaption, caption}
\usepackage{adjustbox}
\renewcommand{\UrlFont}{\ttfamily\small}
\usepackage{mathtools}
\usepackage{wrapfig}
\usepackage{algorithmic}
\usepackage{color}
\usepackage{graphicx}

\newcommand{\RN}[1]{%
	\textup{\lowercase\expandafter{\it \romannumeral#1}}%
}

\usepackage{microtype}
\graphicspath{{figs/}}
\allowdisplaybreaks[1]

\ifx\proof\undefined
\newenvironment{proof}{\par\noindent{\bf Proof\ }}{\hfill\BlackBox\\[2mm]}
\fi

\ifx\theorem\undefined
\newtheorem{theorem}{Theorem}
\fi
\ifx\example\undefined
\newtheorem{example}{Example}
\fi
\ifx\lemma\undefined
\newtheorem{lemma}[theorem]{Lemma}
\fi

\ifx\corollary\undefined
\newtheorem{corollary}[theorem]{Corollary}
\fi

\ifx\assumption\undefined
\newtheorem{assumption}{Assumption}
\fi

\ifx\definition\undefined
\newtheorem{definition}{Definition}
\fi

\ifx\proposition\undefined
\newtheorem{proposition}[theorem]{Proposition}
\fi

\ifx\remark\undefined
\newtheorem{remark}{Remark}
\fi

\ifx\conjecture\undefined
\newtheorem{conjecture}[theorem]{Conjecture}
\fi

\ifx\factoid\undefined
\newtheorem{factoid}[theorem]{Fact}
\fi

\ifx\axiom\undefined
\newtheorem{axiom}[theorem]{Axiom}
\fi

\graphicspath{{figures/}}

\makeatletter
\newcommand*{\rom}[1]{\expandafter\@slowromancap\romannumeral #1@}
\makeatother

\input{Definitions}

\def\RL{\textsf{RL}}
\def\IRL{\textsf{IRL}}

\ifx\remark\undefined
\newtheorem{remark}{Remark}
\fi

\definecolor{tblue}{HTML}{174992}

\title{SDA: Improving Text Generation with Self Data Augmentation}
\author{
    Ping Yu\textsuperscript{\rm 1},
    Ruiyi Zhang\textsuperscript{\rm 2},
    Yang Zhao\textsuperscript{\rm 1},
    Yizhe Zhang\textsuperscript{\rm 3},
    Chunyuan Li\textsuperscript{\rm 3},
    Changyou Chen\textsuperscript{\rm 1}\\
}
\affiliations{
    \textsuperscript{\rm 1} The State University of New York at Buffalo \\
    \textsuperscript{\rm 2} Duke University \\
    \textsuperscript{\rm 3} Microsoft Research, Redmond \\
    pingyu@buffalo.edu

}
\begin{document}

\maketitle

% \linenumbers  %

\begin{abstract}
Data augmentation has been widely used to improve deep neural networks in many research fields, such as computer vision. However, less work has been done in the context of text, partially due to its discrete nature and the complexity of natural languages. In this paper, we propose to improve the standard maximum likelihood estimation (MLE) paradigm by incorporating a self-imitation-learning phase for automatic data augmentation. 
Unlike most existing sentence-level augmentation strategies, which are only applied to specific models, our method is more general and could be easily adapted to any MLE-based training procedure. 
In addition, our framework allows task-specific evaluation metrics to be designed to flexibly control the generated sentences, for example, in terms of controlling vocabulary usage and avoiding nontrivial repetitions. Extensive experimental results demonstrate the superiority of our method on two synthetic and several standard real datasets, significantly improving related baselines.

\end{abstract}

%###############################################################################
%###############################################################################
\section{Introduction}

Deep learning models are often considered data-hungry, in the sense that the more data we have, the better performance we can achieve. In practice, it is usually too laborious and time-consuming to annotate a large amount of training data. An appropriate data augmentation method can significantly boost the performance of a deep-learning model. Generally speaking, augmenting continuous data such as images is relatively simpler, as one could easily infer the meaning of an image even if it is manipulated by flipping, adding noise, resizing, {\it etc}. However, for discrete data such as those in the natural-language-processing (NLP) filed, data augmentation is more challenging. This is partly because even a single-word change could affect the meaning of a whole sentence.

Recent advances of NLP has advocated to adopt pre-trained models to effectively leverage extensive unlabeled data collected from the internet. Such a learning paradigm has achieved defeating success for generating human-level sentences and articles. Nevertheless, we argue that a large amount of data still benefits the finetuning stage, which can be achieved by data augmentation. In fact, data augmentation for text is not new. The traditional way is to replace several words in a sentence with its synonyms (words or phrases having similar meanings) \cite{zhang2015character,wang2015s,kobayashi2018contextual}. However, the effect of such word-level replacement is limited. A better option is to use sentence-level augmentation, which has been successfully used in many applications. For example, \cite{sennrich2015improving,sugiyama2019data} used the back-translation method to generate more training data to improve neural machine translation (NMT); \cite{kafle2017data} generated question and answering pairs using LSTM for the Visual Question Answering (VQA) task. A noticeable drawback of these sentence-level text augmentation methods, however, is the lack of universality in the sense that they are only applicable to specific models. 

In this paper, we overcome this limitation by proposing a self-imitation-learning based method to generate augmented text data. In general, our method contains two learning phases:
\begin{itemize}
    \item[$\RN{1})$] {\em The MLE Training stage}: our model starts from the regular MLE based training until convergence;
    \item[$\RN{2})$] {\em The Self Data Augmentation stage}: this phase maintains a buffer $\mathcal{D}_B$ to store top-k ranked generated data samples as the {\it self-augmented data}. In training, it also updates a discriminator to alternatively match the model generated data with the {\it self-augmented data} and the training data. 
    %We further derive the formula of {\it Inverse Reinforcement Learning} and divide the initial steps into two sub-steps:
    %the generator imitates training and self-augmented data alternatively under the guidance of a reward function provided by the discriminator. 
    % \cy{Why does the generator need to imitate training and self-augmented data alternatively? I think this is the key difference with other ideas and worth highlighting, and the reason can be explained formally with the SEIRL formulation. But some intuitive explanations are needed here.}
\end{itemize}
Especially, the rationality of our self-data-augmentation stage is supported by a proposed {\it self-inverse reinforcement-learning framework}, which reveals that the alternative matching in the discriminator leads to learning a policy that prefers generating higher-quality text. 
%Essentially, in the second stage, our model is designed to imitate training data as well as {\it self-augmented data} to further improve the MLE training model.  
The major contributions of this work are summarized in three aspects:
\begin{itemize}
    \item We propose a flexible sentence-level text augmentation technique that could be combined with any MLE based text generation method to improve model performance. 
    \item We show how to control the generated data by defining task-specific evaluation metrics to rank the {\it self-augmented data} in the buffer, {\it e.g.}, controlling vocabulary usage and avoiding nontrivial repetition.
    \item We evaluate our model on both unconditional generation and conditional generation tasks, achieving significant improvements relative to the baselines. We further conduct human evaluations, grammar check evaluations, and memorization evaluations to demonstrate our method's superiority. Ablation studies are conducted to investigate the effects of varying training set size and removing the imitation-learning component.

    % Moreover, we discuss the ability of our model to control vocabulary usage and avoid nontrivial repetition in generation.
\end{itemize}

\section{Preliminaries and Related Works}
% \vspace{-0.2cm}
\subsection{Text augmentation}

The traditional way of text augmentation mainly focuses on word-level cases. \citet{zhang2015character} proposed to select a word according to the geometric distribution and replace it with its synonym. \citet{wang2015s} found synonyms with k-nearest neighbors using word embedding. \citet{kobayashi2018contextual} proposed to replace a word by the output of a bidirectional RNN language model. EDA \cite{wei2019eda} proposed four operations: synonym replacement, random insertion, random swap, and random deletion. As acknowledged, the performance of word-level text augmentation is limited due to simple word replacement in a sentence without considering the sentence structure. 

There have been some attempts for sentence-level text augmentation, but they were only designed for specific applications. In a translation task, \citet{sennrich2015improving} designed back translation. \citet{fadaee2017data} generated new sentence pairs containing rare words in new and synthetically created contexts. \citet{zheng2019mirror} designed a mirror architecture to augment translation pairs for non-parallel data. For the Visual Question Answering (VQA) task, \citet{kafle2017data} generated question-answer pairs using LSTM. It is worth noting that the aforementioned methods are not universal in terms of models. 

By contrast, our framework a universal sentence-level method for text augmentation, which could be easily combined with the popular MLE based methods for training. Besides, our augmentation process can be easily controlled ({\it e.g.}, control the vocabulary usage and avoid nontrivial repetitions in the generation) by incorporating some user-defined metrics. 
% We compare our model with the EDA method \cite{wei2019eda} for text augmentation.

% \vspace{-0.2cm}
\subsection{Text generation as reinforcement learning}
Text generation can be explained as a reinforcement-learning problem. To generate a text sequence, neural language models~\cite{mikolov2010recurrent} typically adopt an auto-regressive manner. 
Given a dataset $\mathcal{D}=\left\{\left({y_{1:T}}^{(i)}\right)\right\}$. Denote ${y_{1:t-1}}^{(i)}$ as all tokens before token $t$. $T$ is the length of the sentence ${\mathbf{y}}^{(i)}$. The standard MLE approach to train the model $p_{\theta}$ minimizes:
\begin{small}
\begin{equation*}
    \hspace{-0.1cm}\mathcal{L}_{\mathrm{MLE}}\left(p_{\theta},{\mathbf{y}}^{(i)}\right)=-\sum_{t=1}^{T} \log p_{\theta}\left({y_{t}}^{(i)} | {y_{1:t-1}}^{(i)}\right)
\end{equation*}
\end{small}
The above sequential text generation process has been reformulated as reinforcement learning in SeqGAN \cite{YuZWY:AAAI17}. In the following, we will omit the sentence index ``$i$'' when the context is clear. 
In the timestep $t$, the state $s$ is the current produced tokens $y_{1:t-1}$ and the action $a$ is the next token $y_{t}$ to be selected. The policy model (or generator), denoted as $G_{\theta}\left(y_{t} | y_{1: t-1}\right)$, is usually stochastic and defines the probability of generating the next token given previous ones. The state transition is deterministic after an action has been chosen, {\it i.e.}, $\delta_{s, s^{\prime}}^{a}=1$ for the next state $s^{\prime}=y_{1: t}$ if the current state $s=y_{1: t-1}$ and the action $a=y_{t}$; and $\delta_{s, s^{\prime \prime}}^{a}=0$ for other next states $s^{\prime \prime}$. In these methods, usually a discriminator network, denoted as $D_{\phi}$, is learned to provide the reward for the REINFORCE algorithm \cite{williams1992simple}. The discriminator could only calculate a reward for a complete sentences instead of short-term reward, which will be a problem for long-text generation. \citet{YuZWY:AAAI17} designed the \textit{N}-Time Monte Carlo search ($\mathrm{MC}$) to calculate a roll out reward.
Specifically, they applied the Monte Carlo search with a roll-out policy $G_{\theta}$ to sample the unknown last $T-t$ tokens. They represent an $N$-time Monte Carlo search as:
\begin{small}
% \small\begin{equation}
\begin{equation}
    \left\{{y_{1: T}}^{(1)}, \ldots, {y_{1: T}}^{(N)}\right\}=\mathrm{M} \mathrm{C}^{G_{\theta}}\left(y_{1: t} ; N\right)~,
\label{eq:reward_1}
\end{equation}
\end{small}
where ${y_{1: t}}=\left(y_{1}, \ldots, y_{t}\right)$ and ${y_{t+1: T}}^{(n)}$ is sampled based on the generator policy $G_{\theta}$ and the current state. 
To reduce the variance and get a more accurate assessment of the action value, they ran the roll-out policy starting from the current state until the end of the sequence for $N$ times to get a batch of output samples. The roll-out reward $Q_{D_{\phi}}^{G_{\theta}}$ is calculated as
% \cy{$Q$ and $D$ undefined}:
% \vspace{-0.3cm}
% $Q_{D_{\phi}}^{G_{\theta}}\left(s=Y_{1: t-1}, a=y_{t}\right)=$
% \[
% \left\{\begin{array}{ll}
% \frac{1}{N} \sum_{n=1}^{N} D_{\phi}\left(Y_{1: T}^{n}\right), Y_{1: T}^{n} \in \mathrm{MC}^{G_{\beta}}\left(Y_{1: t} ; N\right) & \text { for } t<T \\
% D_{\phi}\left(Y_{1: t}\right) & \text { for } t=T
% \end{array}\right.
% \]
\begin{small}
\begin{align}
\label{eq:reward_2}
& Q_{D_{\phi}}^{G_{\theta}}\left(s=y_{1: t-1}, a=y_{t}\right) =  \\ \nonumber
& 
\left\{\begin{array}{ll}
\frac{1}{N} \sum_{n=1}^{N} D_{\phi}\left({y_{1: T}}^{(n)}\right), & \text { for } t<T \\
\\
D_{\phi}\left(y_{1: t}\right) & \text { for } t=T
\end{array}\right.
\end{align}
\end{small}
The objective function w.r.t. the generator parameters $\theta$ is
%\vspace{-0.3cm}
\begin{small}
\begin{align}
\label{eq:rl}
J(\theta) = \sum_{t=1}^{T} \mathbb{E}_{y_{t} \sim G_{\theta}\left(y_{t} | y_{1: t-1}\right)}[Q_{D_{\phi}}^{G_{\theta}}(y_{1: t-1}, y_{t})]
\end{align}
\end{small}
The gradient is calculated through policy gradient:
\begin{small}
\begin{align}
\label{eq:policy_gradient}
\nabla_{\theta}J(\theta) &=\sum_{t=1}^{T} \mathbb{E}_{y_{t} \sim G_{\theta}\left(y_{t} | y_{1: t-1}\right)}  \\ \nonumber
&\left[\nabla_{\theta} \log G_{\theta}\left(y_{t} | y_{1: t-1}\right) \cdot Q_{D_{\phi}}^{G_{\theta}}\left(y_{1: t-1}, y_{t}\right)\right]
\end{align}
\end{small}
LeakGAN \cite{GuoLCZYW:AAAI18} improved the SeqGAN framework through leaking feature information from discriminator to generator and treating it as additional guided information for training generator. 
% More details are presented in Appendix~\ref{app:preliminaries IRL}.

%###############################################################################
\section{Self-Improved Text Augmentation}
\label{sec:method}
Imitation learning is a framework of training policy to imitate training data. The recently developed self imitation learning \cite{oh2018self} explores the possibility of using the generated data to improve a model further. In this paper, we leverage the advances of these ideas and propose a model to use self-augmented data to improve MLE-based models for data augmentation. 

Specifically, in text generation, we treat an MLE-based model, $p_{\theta}\left(y_{t} | y_{1:t-1}\right)$, as a generator network $G_{\theta}\left(y_{t} | y_{1: t-1}\right)$, which aims to learn a policy to generate next word based on previous word. Some example generators can be LSTM-based models, CNN-based models and Transformer-based models. On top of this, we add a discriminator network $D_{\phi}$ to provide reward to be applicable to the reinforcement-learning setting. 
Based on this setting, our self data-augmentation method consists of two stages in training. The first stage is the general MLE-based model training (\textit{MLE Training stage}); and the second stage further refines the MLE-based model with some self-augmented data, called the \textit{Self Data Augmentation stage}. Note for the general MLE-based model, we typically add the second stage after the convergence of the MLE training to further improve the MLE-based model. Below we explain these two stages in detail, with a formal explanation of the second stage given in the next section.
% \cy{Again, why these two stages? What does each stage achieve?}

\subsection{The MLE Training stage}
In this stage, the generator $G_{\theta}\left(y_{t} | y_{1: t-1}\right)$ and the discriminator $D_{\phi}$ are trained separately. The generator is trained with the general MLE training mechanism, and the discriminator is trained in a GAN frame \cite{GoodfellowAMXWOCB:NIPS14}. Specifically, the generator is trained with an objective:
\begin{small}
\begin{equation}
\label{eq:stage1_g}
    J_g(\pi)= \sum_{t=1}^{T} \log G_{\theta}\left(y_{t} | y_{1:t-1}\right).
\end{equation}
\end{small}
% \vspace{-1mm}
The discriminator performs as a classifier to discriminate the generated sentences and ground truth training sentences with an objective:
\begin{small}
\begin{align}\label{eq:stage1_d}
J_d & = \mathbb{E}_{y_{1: T} \sim p_{\text {data }}}\left[\log D_{\phi}(y_{1: T})\right] \\ \nonumber
& +\mathbb{E}_{y_{1: T} \sim G_{\theta}}\left[\log \left(1-D_{\phi}(y_{1: T})\right)\right]~,
\end{align}
\end{small}
where $p_{\text {data}}$ represents the training data distribution; and $G_{\theta}$ represents the generated-sentence distribution. 
In a large-model setting, one can choose to use a pre-trained MLE-based model and only train the discriminator with its generated sentences and the training data.

% through equation~\ref{eq:mle}. Using generated sentences from generator and training sentences to train the discriminator.

\subsection{The Self Data Augmentation stage}

% The overall procedure of imitation learning stage is shown in Fig.~\ref{fig:seirl}. 

Despite the simplicity, an MLE-trained model usually generates low-quality text in practice. A natural idea for improvement is to select some high-quality sentences from the generated data as the augmented data to enlarge the original training dataset. However, we note two issues with this approach: 
$\RN{1})$ It is hard to control the percentage of augmented data mixed with the training data. $\RN{2})$ If we direct mix augmented data with training data, it might have an adverse effect as there might be some low-quality data in the augmented data.  
To overcome these issues, we propose to separate the augmented and training data by using a buffer $\mathcal{D}_B$ to store the {\it self-augmented data}. 
In this way, $\RN{1})$ we could conveniently control the percentage of augmented data at any time; and $\RN{2})$ the low-quality augmented data will not have a direct effect for the training data.

To apply the imitation-learning framework in our setting, we need to address the issue that there is only one data source in imitation learning. If we were to mix parts of the training data and {\it self-augmented data} into a minibatch at every epoch, we found the training usually unstable because the data in one small minibatch are not representative enough. To overcome this, we propose to divide the {\em self Data Augmentation stage} into two sub-steps, which alternatively updates the discriminator using the training data (called the imitating training data step) and the {\it self-augmented data} in the buffer $\mathcal{D}_B$ (called the imitating augmented data step), respectively. 

% To distinguish between the training dataset and the {\it self-augmented data}, we additionally update the discriminator used in the {\it MLE Training stage}, which gives rewards to the generator trained on the {\it self-augmented data}. In training, we divide the {\em Self Data Augmentation stage} into two sub-steps, which update the discriminator using the training data (the imitating training data step) and the augmented data in buffer $\mathcal{D}_B$ (the imitating augmented data step). 

\begin{remark}
Although these two steps in the {\it Self Data Augmentation stage} seem heuristic, we show that they are actually derived from a more principled view of self-imitation learning framework, which is detailed in the next section. 
\end{remark}

Before describing the two steps, we first define a reference metric used to select generated sentences to construct the {\it self-augmented data} in the buffer $\mathcal{D}_B$. Some examples of the specific metric are given in the experiment part.

\begin{definition}[Reference Metric]
    A reference metric is defined as a function,  $m:\mathcal{S}\rightarrow \mathbb{R}$, mapping a sentence $\sbb \in \mathcal{S}$ to a real value measuring the goodness of the sentence. 
\end{definition}

\vspace{-4mm}
\paragraph{The imitating self-augmented data step}
In this step, we first update the buffer $\mathcal{D}_B$ with the generated sentences associated with the highest reference metric values. Specifically, we associate each sentence with a reference metric score in the buffer. When new data are generated, we replace the low-score sentences in the buffer with the high-score generated sentences, and only keep $k$ highest-score sentences, where $k$ is a hyperparameter depending on the size of the training set and complexity of the task. After this, we assign the {\it self-augmented data} with label 1, and the current generated data with label $0$, which are then used to update the discriminator with the objective \eqref{eq:stage1_d} (but replacing the data distribution $p_{\text {data}}$ with {\it self-augmented data} distribution $p_{\text {aug}}$ for this setting).

As mentioned above, the discriminator provides rewards to the generator to measure the current generated data quality. If a generated sentence is similar to sentences in the buffer $\mathcal{D}_B$, the discriminator will output a larger value. In this way, a large reward will be fed into the generator, which is then updated in a reinforcement-learning setting. Specifically, given the current generated sentence $y_{1:T} \in G_{\theta}$, a reward $Q_{D_{\phi}}^{G_{\theta}}$ is first calculated by \eqref{eq:reward_1} and \eqref{eq:reward_2}. Finally, the generator is updated with the following gradient: $\nabla_{\theta}J(\theta)$
\begin{align}
    \label{eq:stage2_g}
     \hspace{-0.3cm} =\sum_{t=1}^{T} \mathbb{E}_{y_{t} \sim G_{\theta}\left(y_{t} | y_{1: t-1}\right)}\left[\nabla_{\theta} \log G_{\theta}\left(y_{t} | y_{1: t-1}\right) \cdot Q_{D_{\phi}}^{G_{\theta}}\right]
\end{align}

\vspace{-0.5cm}
\paragraph{The imitating training data step}
This step is the same as the previous step except that the data fed to the discriminator is the training data with label $1$ (instead of the {\it self-augmented data}) and the currently generated data with label $0$ through \eqref{eq:stage1_d}. After training the discriminator, the reward is calculate through \eqref{eq:reward_1} and \eqref{eq:reward_2}, which are then used to update the generator with the objective \eqref{eq:stage2_g}. Replacing the augmented data in the {\em imitating self-augmented data step} with training data would make the generator imitate the training data.

\begin{small}
    \begin{algorithm}
    	\caption{Self Data Augmentation}
    	\label{alg:seirl}
    	\begin{algorithmic}[1]
    		\REQUIRE
    		$\RN{1})$ Generator network $G_{\thetab}$; 
    		$\RN{2})$ Discriminator network $D_{\phi}$;
    		$\RN{3})$ Training dataset $\mathcal{D}_E $; 
    		$\RN{4})$ Reference metric $m(\sbb)$; 
    		$\RN{5})$ the buffer $\mathcal{D}_B$ to store {\it self-augmented data}. 

    		\STATE Initialize $\pi_{\thetab}$, $r_{\phib}$; initialize the buffer $\mathcal{D}_B \leftarrow \emptyset$
    % 		, and $\mathcal{B}_h$ with randomly generated sentences.
            \STATE \footnotesize{\color{blue} \# Stage I: {\em MLE Training stage}}
            \REPEAT
            \STATE Train generator $G_{\theta}$ via MLE with (\ref{eq:stage1_g})
            \STATE Train discriminator $D_{\phi}$ with (\ref{eq:stage1_d})
            \UNTIL{converge}
            \STATE  \footnotesize{\color{blue} \# Stage II: {\em Self Data Augmentation stage}}
            % \STATE  \footnotesize{\textcolor{tblue} {\# Stage II: {\em Self Data Augmentation stage}}}
    		\REPEAT 
    		
    % 		\STATE \footnotesize{\color{blue} \# Replay-Buffer Update}
    		
    		\STATE{Generate a set of sentences $\Wb$ from policy $G_{\theta}$, Update the buffer $\mathcal{D}_B$ by choosing the sentences from $\Wb$ with highest reference-metric values.}

    		\IF{{\em Imitating self augmented data step}}
    		\STATE{Sample sentences $\Sbb \sim \pi_{\theta} $ and $\Sbb_B \sim \mathcal{D}_B $} 
            \STATE{Update the discriminator $D_{\phi}$ using $\Sbb$ and $\Sbb_B$ with (\ref{eq:stage1_d}})
    		\ELSIF{{\em Imitating training data step}}
    % 		\STATE \footnotesize{\color{blue} \# step 2}
    		\STATE{Sample sentences $\Sbb \sim \pi_{\theta} $ and $\Sbb_E \sim \mathcal{D}_E $}
    		\STATE{Update the discriminator $D_{\phi}$ using $\Sbb$ and $\Sbb_E$ with (\ref{eq:stage1_d})}
    		\ENDIF
    % 		\vspace{0.2cm}
    		\STATE{Update the reward $Q_{D_{\phi}}^{G_{\theta}}$ with (\ref{eq:reward_1} and \ref{eq:reward_2})}
    		\STATE{Update the generator $G_{\theta}$ with (\ref{eq:stage2_g})}
    		
    		\UNTIL{Converge}
    	\end{algorithmic}
    \end{algorithm}
    \vspace{-2mm}
\end{small}
% \vspace{-0.4cm}

%\vspace{-0.2cm}
\subsection{The whole process}
To summarize, the {\em Self Data Augmentation stage} contains two steps: the {\em imitating self augmented data step} imitates the self-augmented data, while the {\em imitating training data step} imitates the training data. These two steps run alternately. In the beginning, we propose to run more {\em imitating training data steps} and gradually increase the {\em imitating self augmented data step}. In our experiment, we run six {\em imitating training data steps} at the beginning. After updating the discriminator and generator, we run one {\em imitating self augmented data step}. After every ten epochs, we decrease one step of the {\em imitating training data steps} until it reduces to one. 
The whole algorithm is described in algorithm~\ref{alg:seirl}.

%###############################################################################
\vspace{-2mm}
\section{Understanding the Self Data Augmentation Stage as IRL with Self Enhancement}
% \iris{AAAI template cannot refer sections}
% \cy{Previous sections should refer to this section to explain why the self data augmentation stage is defined that way.} 
In this section, we show that the {\em self data augmentation stage}, which consists of alternatively updating the discriminator with the two data sources, can be explained as a procedure to solve an inverse reinforcement learning (IRL) problem with self-enhancement. 
% To this end, note IRL aims to learn an optimal reward function from the {\em expert trajectories} $\mathcal{D}_E$, which is represented as a training dataset in text generation. Typically, $\mathcal{D}_E$ consists of state-action pairs generated from an expert policy $\pi_E$. 
The goal of IRL is to recover the optimal reward function $r^*(\cdot, \cdot)$ as well as the corresponding policy $\pi^*$ \cite{NIPS2016_6391}:
\begin{small}
{\small\begin{align}\label{eq:IRL_main}
	&\{r^*, \pi^*\} = \arg \min_{\pi\in\Pi}\max_{r\in\mathbb{R}^{\mathcal{S}\times \mathcal{A}}} \mathcal{L}(\pi, r)  - \psi(r)\\ 
	&= 
	 \arg\max_{r\in\mathbb{R}^{\mathcal{S}\times \mathcal{A}}} \sum_{\sbb, \ab}\rho_{E}(\sbb, \ab)r(\sbb, \ab) \nonumber\\ 
	 & -[\max_{\pi\in\Pi}H(\pi) + \sum_{\sbb, \ab}\rho(\sbb, \ab)r(\sbb, \ab)] - \psi(r) \nonumber,
% 	&= -H(\pi)\! +\! \sum_{\sbb, \ab}\rho_{E}(\sbb, \ab)r(\sbb, \ab)
% 	\!-\! \sum_{\sbb, \ab}\rho(\sbb, \ab)r(\sbb, \ab) - \psi(r)~ \nonumber,
\end{align}}
\end{small}
% \begin{small}
% {\small\begin{align*}\label{eq:IRL_main}
% 	&\{r^*, \pi^*\} \triangleq \IRL(\pi_E)\\ 
% 	&= 
% 	 \arg\max_{r\in\mathbb{R}^{\mathcal{S}\times \mathcal{A}}} \sum_{\sbb, \ab}\rho_{E}(\sbb, \ab)r(\sbb, \ab)\\ 
% 	 & -[\max_{\pi\in\Pi}H(\pi) + \sum_{\sbb, \ab}\rho(\sbb, \ab)r(\sbb, \ab)] \\
% 	& = \arg\max_{r\in\mathbb{R}^{\mathcal{S}\times \mathcal{A}}} \min_{\pi\in\Pi} ~~\mathcal{L}(\pi, r)\\
% 	& \triangleq \! -H(\pi)\! +\! \sum_{\sbb, \ab}\rho_{E}(\sbb, \ab)r(\sbb, \ab)
% 	\!-\! \sum_{\sbb, \ab}\rho(\sbb, \ab)r(\sbb, \ab)~,
% \end{align*}}
% \end{small}
% \vspace{-1mm}
where $\rho(\sbb, \ab)$ is the {\em occupancy measure} defined as the stationary joint distribution of $(\sbb, \ab)$ under policy $\pi$, $\psi(\cdot)$ is the regularizer. 

% We note this setting is sub-optimal if the given data is noisy, as the learned policy is expected to perform worse than the expert policy.
% % (and equal to the expert policy at optimality). 
% Besides, the IRL objective is ill-defined, which often induces multiple valid solutions to the problem because the solution space is too flexible \cite{Ng:2000:AIR:645529.657801,Finn:2016:GCL:3045390.3045397}. To address these problems, some constraints are typically placed on the reward functions. Finally, the IRL is formulated with a regularizer $\psi(\cdot)$ as
% \begin{small}
% \begin{align}\label{eq:IRL_main}
% \{\pi^*, r^*\} =& \arg \min_{\pi\in\Pi}\max_{r\in\mathbb{R}^{\mathcal{S}\times \mathcal{A}}} 
% \mathcal{L}(\pi, r)  - \psi(r).
% \end{align}\par 
% \end{small}

The traditional regularization in IRL only considers constraining the model capacity of the policy~\cite{ho2016generative}, which makes it hard to be adapted to diverse tasks. 
Natural languages own intrinsic characteristics based on syntax and semantics. 
To leverage this prior knowledge, we propose incorporating user-defined evaluation measures into $\psi(r)$ so that the learned policy aligns well with the pre-defined metrics when generating unseen sentences. 
Specifically, we define $\psi(r)$ in \eqref{eq:IRL_main} to be a mixture of a {\it capacity regularizer}  $\psi_1(r)$ and a {\it self-guided regularizer}  $\psi_0(r)$:
\begin{small}
\begin{align}\label{eq_mix_regularier_main}
	\psi(r) = -\lambda \psi_0(r) + \psi_1(r)~, % 
\end{align}
\end{small}
with $\lambda > 0$ the hyper-parameter to balance two terms. 

\vspace{-0.2cm}
\paragraph{Capacity regularizer}~
It has been shown in~\cite{ho2016generative} that the adversarial regularizer can provide better generalization.  
To imitate the expert policy, we adopt the {\it capacity regularizer}  $\psi_1(r)$ in GAIL~\cite{ho2016generative}:
The specific form is given in the appendix \ref{app:IL}. 
% \cy{Write it out here if there is enough space.}
% \begin{align}\label{eq_capacity_regularizer}
% 	\psi_1(r) \triangleq \left\{
% 		\begin{array}{ll}
% 			\mathbb{E}_{\pi_E}\left[g(r(\sbb, \ab))\right] & \mbox{if } r(\sbb, \ab) \geq 0\\
% 			+\infty & \mbox{otherwise}
% 		\end{array}~,
% 	\right.
% \end{align}
% where $g(x)=\left\{\begin{array}{ll}-x-\log \left(1-e^{x}\right) & \text { if } x<0 \\ +\infty & \text { otherwise }\end{array}\right.$

\vspace{-0.2cm}
\paragraph{Self-guided regularizer}~To promote generating suboptimal demonstrations that are well aligned with the diverse tasks, we propose the {\it self-guided regularizer} as a weighted linear function over rewards: 
\begin{small}
\begin{align}\label{eq_semantic_regularier_main}
	 \psi_0(r) = \sum_{\sbb, \ab}q(\sbb, \ab)r(\sbb, \ab) 
\end{align}\par
\end{small}
\vspace{-1mm}
\noindent where $q(\sbb, \ab)$ is a probability function constructed via evaluation metrics. The specific form of $q$ is defined in Appendix. Generally, this regularizer encourages higher rewards in regions where $q(\sbb, \ab)$ are large.
% In text generation, the evaluation metrics are used to evaluate certain quality aspects of a generated sentence, such as the BLEU scores, self-BLEU scores. As a result, the $q$ value is defined to be larger wherever the BLEU score is highers (or wherever the self-BLEU score is lower).

By plugging the capacity regularizer and the self-guided regularizer into~\eqref{eq:IRL_main}, we formulate our objective as an optimization problem to find the optimal reward $r^*$ and the optimal policy $\pi^*$:
{\small
\begin{align}\label{eq:SEIRL_main}
	&\{\pi^*, r^*\} = \arg\min_{\pi\in\Pi}\max_{r\in\mathbb{R}^{\mathcal{S}\times \mathcal{A}}} -2H(\pi)- \psi_1(r)  +   \\
& \!\! \sum_{\sbb, \ab}q(\sbb, \ab)r(\sbb, \ab) \!\! + \!\! \sum_{\sbb, \ab}\rho_E(\sbb, \ab)r(\sbb, \ab) \!\! -  \!\! 2\sum_{\sbb, \ab}\rho(\sbb, \ab)r(\sbb, \ab),\nonumber
\end{align}}
% \vspace{-2mm}
\noindent where we have set $\lambda = 1$ in~\eqref{eq_mix_regularier_main}, and the factor ``2'' is introduced for formulation convenience. The problem \eqref{eq:SEIRL_main} is called {\em IRL with self enhancement}.
% To emphasize the distinction of enhancing the expert trajectories under the guidance of reference metrics, we call our method {\it Self-Improved Inverse Reinforcement Learning} (SE-IRL).  

\vspace{-0.2cm}
\paragraph{Solving \eqref{eq:SEIRL_main} with a two-step algorithm}~
Due to the introduction of $q(\sbb, \ab)$, solving \eqref{eq:SEIRL_main} directly is difficult. We reformulate \eqref{eq:SEIRL_main} with an augmented policy $\pi_1$:
% \vspace{-0.4cm}
{\small\begin{align}\label{eq:SEIRL1_main}
	 & \min_{\pi_1, \pi\in\Pi}\max_{r\in\mathbb{R}^{\mathcal{S}\times \mathcal{A}}} -H (\pi_1) + \sum_{\sbb, \ab}(q(\sbb, \ab) - \rho(\sbb, \ab))r(\sbb, \ab)\nonumber \\
	& - H(\pi) - \psi_1(r) + \sum_{\sbb, \ab}(\rho_E(\sbb, \ab) - \rho(\sbb, \ab))r(\sbb, \ab),
	\nonumber\\
	&~~~~~~~~~~~~~~~~~~~~\mbox{s.t. }~~ r_1 = r,~~ \pi_1 = \pi
\end{align}}
% \vspace{-1mm}
Objective \eqref{eq:SEIRL1_main} now allows us to decompose the original problem into two sub-problems. Adopting ideas from ADMM \cite{Boyd:2011:DOS:2185815.2185816}, we decompose \eqref{eq:SEIRL1_main} into two sub-problems, which represent as the objectives of the {\it imitating self augmented data step} (objective \eqref{eq:SEIRL-1_main}) and the {\it imitating training data step} (objective \eqref{eq:SEIRL-2_main}) in our proposed data-augmentation framework:
{\small\begin{flalign}
	&\min_{\pi_1\in\Pi}\max_{r\in\mathbb{R}^{\mathcal{S}\times \mathcal{A}}}\mathcal{L}_1(\pi_1, r_1) 
	\triangleq -H(\pi_1) \label{eq:SEIRL-1_main}\\
	&+ \sum_{\sbb, \ab}(q(\sbb, \ab) - \rho(\sbb, \ab))r(\sbb, \ab) + \lambda W_1(\pi, \pi_1) \nonumber \\
%\end{flalign}}
%\vspace{-0.6cm}
%{\small\begin{flalign}\label{eq:SEIRL-2_main}
    &\min_{\pi\in\Pi}\max_{r\in\mathbb{R}^{\mathcal{S}\times \mathcal{A}}}\mathcal{L}_2(\pi, r) \triangleq - H(\pi) \label{eq:SEIRL-2_main}\\ 
    &- \psi_1(r)+ \sum_{\sbb, \ab}(\rho_E(\sbb, \ab) - \rho(\sbb, \ab))r(\sbb, \ab) 
    + \lambda W_1(\pi, \pi_1)~. \nonumber
%\vspace{-0.4cm}
\end{flalign}}
The regularizer $W_1(\pi, \pi_1)$ denotes the 1-Wasserstein distance, and is introduced to accommodate the constraint of $\pi_1 = \pi$ in the original problem \eqref{eq:SEIRL1_main}. 
The two phases \eqref{eq:SEIRL-1_main} and \eqref{eq:SEIRL-2_main} are solved alternatively. More details are shown in Appendix. 

\vspace{-.2cm}
\section{Experiments}
We conduct experiments on two synthetic datasets and two real datasets for the unconditional generation task. The unconditional generation task means input for the generator $G_{\theta}$ is pure Gaussian noise, and the generated sentences should match the distribution with the training data distribution. We further test our method on a conditional generation task -- unsupervised style transfer, detailed in the Appendix.

% We conduct experiments on two synthetic datasets and several real datasets. More results are presented in Appendix \ref{app:results}.

\vspace{-2mm}
\subsection{Experimental settings}

\paragraph{Model architectures and baselines} Two model bases:
% We adopt two model bases:

$\RN{1})$
Both the generator and discriminator bases are RNN architectures. We test this model base with only the MLE training (denoted as {\em RNN}). For the ablation study, we implement this model base with the {\em Imitation Learning stage} but with only the {\em imitating training data step} without using augmented data, denoted as {\em RNN+RL}. Our method, combined with this model base, is denoted as {\em RNN+SDA}. Since the {\em EDA} \cite{wei2019eda} method could be applied with the general MLE training paradigm, we compare our model with {\em RNN+EDA}. In addition, \cite{li2019don,roller2020recipes} demonstrate the usefulness of the unlikelihood training mechanism for controlling vocabulary usage and nontrivial repetition, which could be easily combined with MLE training models. We thus compare our {\em SDA} method with unlikelihood training, denoted as {\em RNN+UN} for vocabulary usage control experiment and repetition control experiment.

% For controlling generated sentences experiments, we adapt {\em EDA} method \cite{wei2019eda} for comparison, which called {\em RNN+EDA}.

$\RN{2})$
The second model base adapts the ``leak'' architecture from LeakGAN \cite{GuoLCZYW:AAAI18}. The Discriminator $D_{\phi}$ consists of a feature extractor and an output layer (the softmax classification layer), whose output ranges in between $[0, 1]$. The Generator is designed to be a hierarchical RNN network and can use sentence features to generate the next-word. We call this model base with only the MLE training {\em LeakRNN}. We further implement models {\em LeakRNN+RL}, {\em LeakRNN+SDA}, {\em LeakRNN+EDA} and {\em LeakRNN+UN} using the same method as first model base. 

\vspace{-0.3cm}
\paragraph{Datasets}
We use two synthetic datasets and several real datasets, including the COCO image-caption dataset, EMNLP2017 WMT dataset\footnote{{\small http://statmt.org/wmt17/translation-task.html}} for unconditional generation, Yelp review dataset for conditional generation in Appendix.
The two synthetic datasets are generated following \cite{YuZWY:AAAI17,GuoLCZYW:AAAI18}. 
Table \ref{tab:dataset} in Appendix \ref{app:results} summarizes statics of all the datasets used in this paper.

\begin{figure*}[ht!]
    \centering
    \begin{subfigure}[t]{0.26\textwidth}
        \centering
        \includegraphics[height=1.48in]{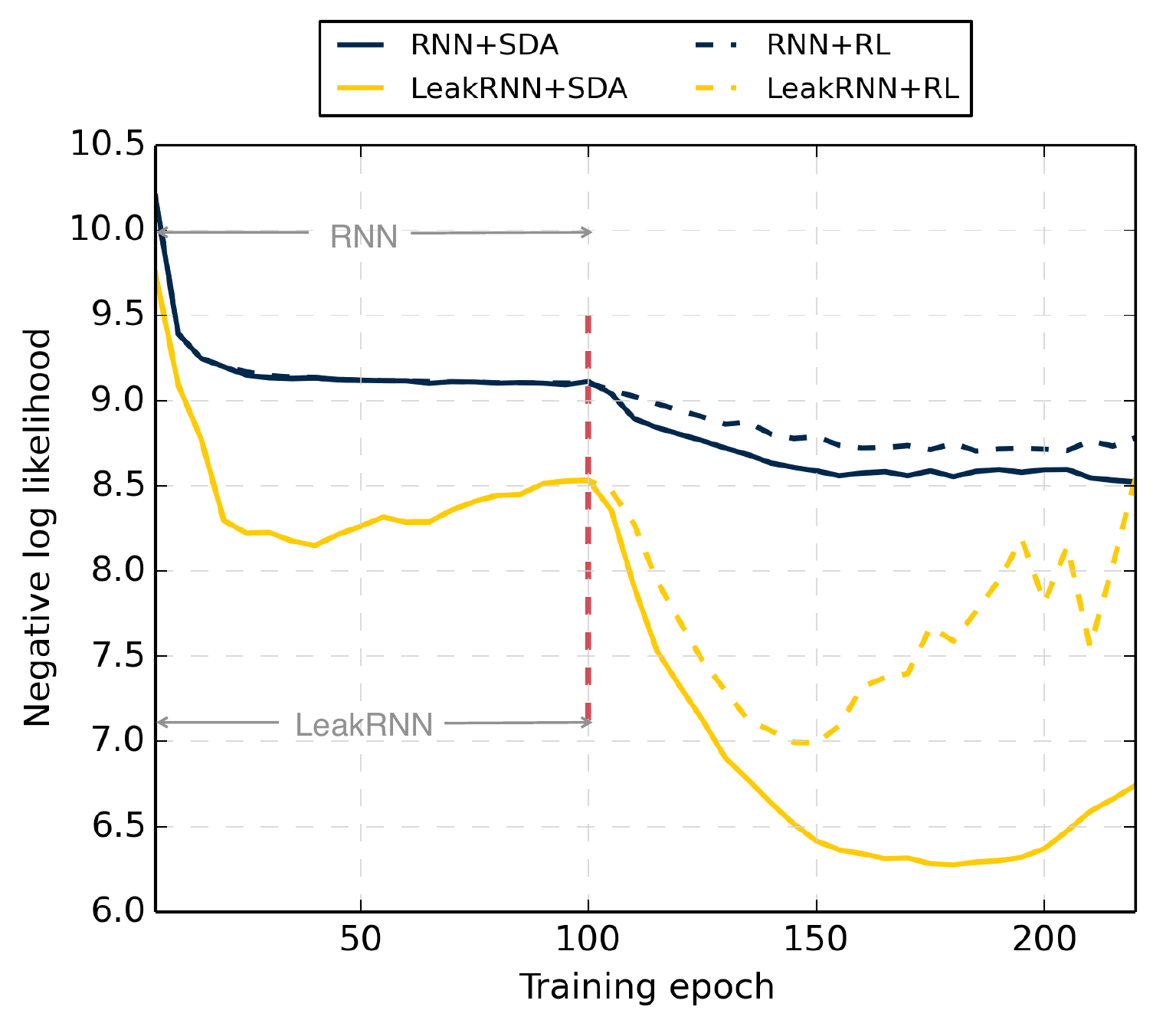}
        \caption{}
        \label{fig:nll20}
    \end{subfigure}%
    \hfill 
    \begin{subfigure}[t]{0.24\textwidth}
        \centering
        \includegraphics[height=1.32in]{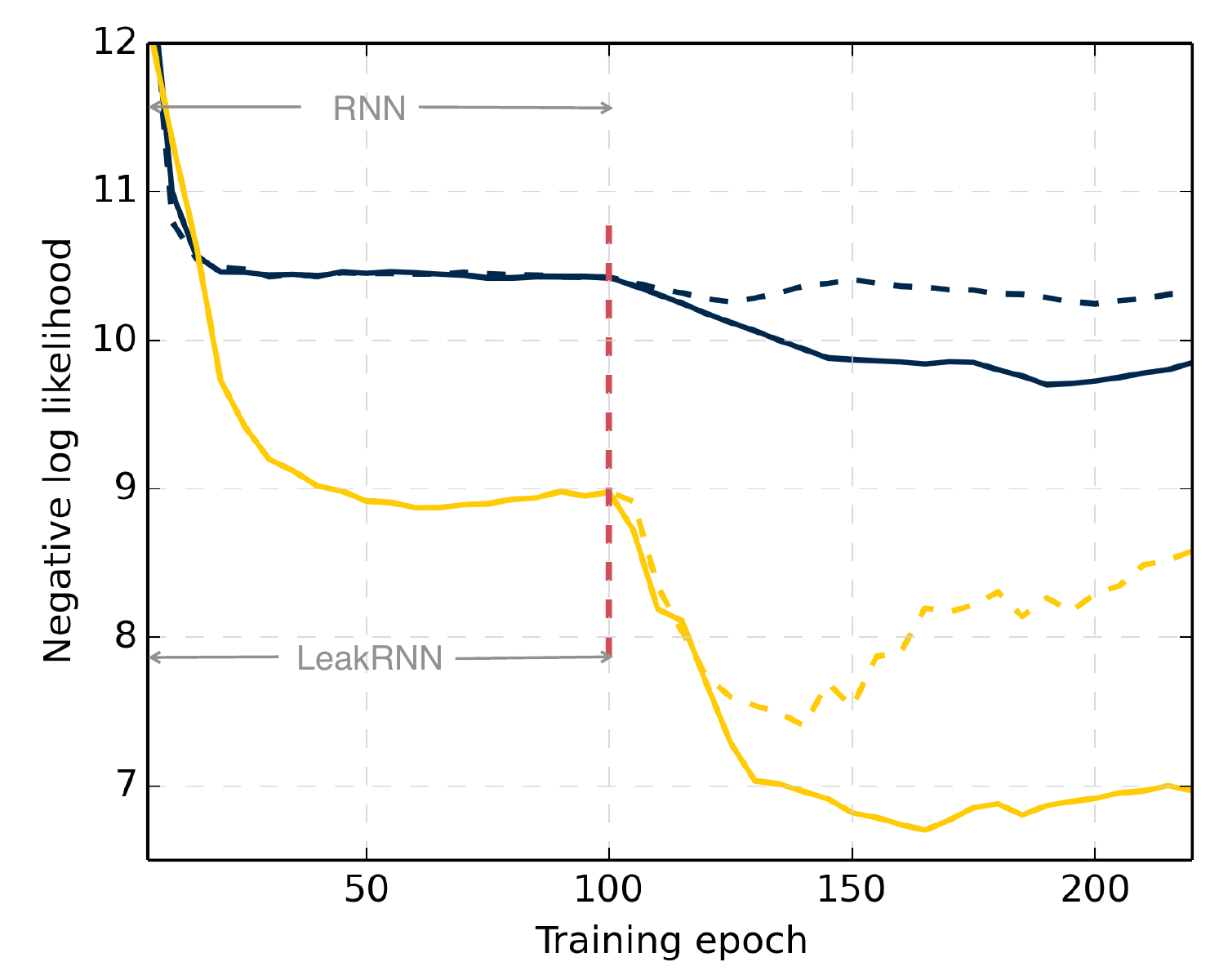}
        \caption{}
        \label{fig:nll40}
    \end{subfigure}
    \hfill
    \begin{subfigure}[t]{0.42\textwidth}
        \centering
        \includegraphics[height=1.32in]{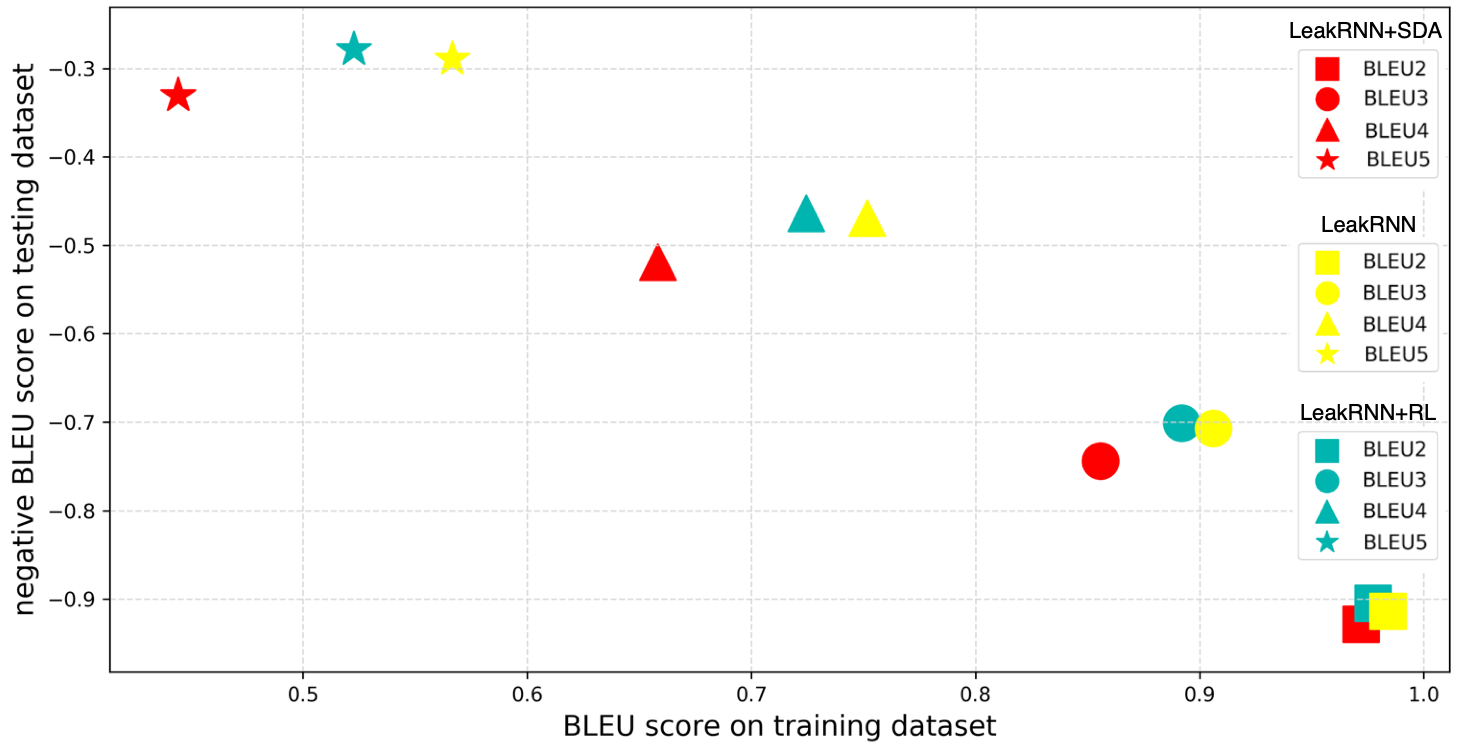}
        \caption{}
        \label{fig:memory}
    \end{subfigure}
    \vspace{-2mm}
    \caption{{\small Training curves on the synthetic dataset with sentence length 20 (a) and length 40 (b). The blue line before 100 epochs is our first model base {\em RNN}. After 100 epochs, the solid blue line is {\em RNN+SDA}, while the blue dotted line is {\em RNN+RL}. The yellow line before 100 epochs is our second model base {\em LeakRNN}. After 100 epochs, the solid yellow line is {\em LeakRNN+SDA}, while the yellow dotted line is {\em LeakRNN+RL}; (c): Test BLEU score between training dataset. A lower value on the y-axis means generative sentences achieve higher accuracy; a Lower value on the x-axis means the higher ability for generating sentences instead of memorializing sentences. }}
    % \vspace{-4mm}
\end{figure*}

\vspace{-2mm}
\subsection{Synthetic-sata experiments}

In order to better verify the effectiveness of our method and track the training process, we adapt the synthetic data experiments followed by \cite{YuZWY:AAAI17,GuoLCZYW:AAAI18}. We leverage a randomly initialized LSTM as an oracle model to generate the real data distribution $p(x_t|x_1,\dots,x_{t-1})$, which is considered to be a human server for real-world problems. Then we use NLL to track the performance of the model during training, defined as
%\vspace{-0.1cm}
\begin{small}
{\begin{align}
\hspace{-0cm}\mathrm{NLL}_{\text {oracle }}=-\mathbb{E}_{y_{1: T} \sim G_{\theta}}\left[\sum_{t=1}^{T} \log G_{\text {oracle }}\left(y_{t} | y_{1: t-1}\right)\right]
\end{align}}
\end{small}
% \vspace{-2mm}
where $G_{\theta}$ and $G_{\text{oracle}}$ denote our generator and oracle network respectively. During training, we leverage $G_{\theta}$ to generate 100,000 sentences and calculate $\mathrm{NLL}_{\text {oracle }}$ every 5 training epoch.

In the experiment, we first run the {\em MLE Training stage} for 100 epochs and store the generated data to the buffer $\mathcal{D}_B$. Then we update our discriminator and generator using the training data and the {\it self augmented data} selected by a reference metric:
%\vspace{-0.1cm}
\begin{small}
\begin{align}\label{eq:reference metrics_nll}
 m(\sbb) \triangleq \mathbb{E}_{y_{1: T} \sim G_{\theta}}\left[\sum_{t=1}^{T} \log G_{\theta}\left(y_{t} | y_{1: t-1}\right)\right]
%  \vspace{-0.7cm}
\end{align}
\end{small}
As we can see from Fig.~\ref{fig:nll20} and \ref{fig:nll40}, before the $100_{th}$ epoch, both {\em RNN} and {\em LeakRNN} models converge. At stage 2, imitation learning (both with and without self augmentation) further improves model performance. However, we observed that {\em RNN+RL} makes the training process less stable.
This phenomenon becomes more obvious as the training process becomes more complicated. 
We suspect that learning a policy for text is more challenging due to the complexity of text itself. Adding our self augmented step enlarges the training dataset, making it easier to learn.

%   #################################################################
\vspace{-2mm}
\begin{table}[h]
\centering
\caption{{\small BLEUs on COCO Image-caption testing dataset.}}
    \vspace{-2mm}
    \begin{adjustbox}{scale=0.7}
    \begin{tabular}{l|cccc}
    \toprule
 & \multicolumn{4}{c}{BLEU $\uparrow$ } \\\ 
 & 2 & 3 & 4 & 5  \\
\midrule
{\em RNN}       & 0.800 & 0.599 & 0.367 & 0.201\\
{\em RNN+RL}     & 0.819 & 0.601 & 0.370 & 0.211\\
{\em RNN+EDA} & 0.822 & 0.610 & 0.373 & 0.220\\
{\em RNN+SDA} (ours) & \textcolor{tblue}{\textbf{0.826}}&\textcolor{tblue}{\textbf{0.612}}&\textcolor{tblue}{\textbf{0.378}}&\textcolor{tblue}{\textbf{0.222}}\\
\hline
{\em LeakRNN}       & 0.916 & 0.789 & 0.594 & 0.404\\
{\em LeakRNN+RL}     & 0.921 & 0.797 & 0.602 & 0.416\\
{\em LeakRNN+EDA} & 0.932 & 0.838 & 0.691 & 0.526\\
{\em LeakRNN+SDA} (ours) & \textcolor{tblue}{\textbf{0.943}}&\textcolor{tblue}{\textbf{0.856}}&\textcolor{tblue}{\textbf{0.738}}&\textcolor{tblue}{\textbf{0.599}}\\
\bottomrule
    \end{tabular}
    \end{adjustbox}
    \label{tab:coco}
\end{table}

\vspace{-4mm}
\subsection{Real data experiments}
% \vspace{-2mm}

For a real data experiment, we run both the {\em MLE Training stage} and the {\em Self Data Augmentation stage} for 100 epochs. At the {\em Imitation Learning stage}, generated sentences are stored in the buffer $\mathcal{D}_B$. The BLEU score \cite{papineni2002bleu} is calculated between sentences in the buffer $\mathcal{D}_B$ and some randomly selected 1,000 samples from the training dataset. We use the BLEU3 score as our reference metric $m(\cdot)$ to construct the buffer. 

According to our observation of real data experiments, directly applying the policy gradient makes the model vulnerable and unstable, which is consistent with synthetic experiment results. 
The time we conduct the test has a significant impact on results. Nevertheless, monitoring the BLEU score in real-time is time-consuming and laborious due to the large testing dataset. Therefore, we examine the BLEU score every ten epochs in the {\em imitation Learning stage} and pick the best results for {\em RNN+RL} and {\em LeakRNN+RL}. Since our method is stable, we only test the BLEU score result after 100 epochs for {\em RNN+SDA} and {\em LeakRNN+SDA}. In addition to the results presented below, we also put some example-sentences in Appendix.
% ~\ref{app:results}.

\vspace{-2mm}
\subsubsection{Short-text generation}
The original COCO dataset consists of a total of 20,734 words and 417,126 sentences. Considering the large gap between the average-length (11) and the max-length (50) of the sentences in this dataset, we only randomly selected sentences with lengths less than 32, ending up with a training dataset of 120,000 sentences and a test dataset of 10,000 sentences. The experiment results are shown in Table~\ref{tab:coco}. Although we do not pick the best result during the training epoch, our self augmentation method ({\em SDA}) still surpasses other baselines in all metrics, demonstrating the effectiveness of our data-augmentation scheme.

%   #################################################################
\vspace{-2mm}
\begin{table}[h]
  \centering
\caption{{\small BLEU score on test data set of EMNLP2017.}}
\vspace{-2mm}
    \begin{adjustbox}{scale=0.7}
    \begin{tabular}{l|cccc}
    \toprule
 & \multicolumn{4}{c}{BLEU $\uparrow$ } \\\ 
 & 2 & 3 & 4 & 5  \\
\midrule
{\em RNN}       & 0.626 & 0.350 & 0.160 & 0.085\\
{\em RNN+RL}     & 0.630 & 0.354 & 0.164 & 0.087\\
{\em RNN+EDA} & 0.639 & 0.361 & 0.172 & 0.106\\
{\em RNN+SDA} & \textcolor{tblue}{\textbf{0.641}}&\textcolor{tblue}{\textbf{0.368}}&\textcolor{tblue}{\textbf{0.176}}&\textcolor{tblue}{\textbf{0.109}}\\
\hline
{\em LeakRNN}       & 0.914 & 0.692 & 0.461 & 0.280\\
{\em LeakRNN+RL}     & 0.916 & 0.707 & 0.469 & 0.289\\
{\em LeakRNN+EDA} & 0.921 & 0.738 & 0.509 & 0.312\\
{\em LeakRNN+SDA} & \textcolor{tblue}{\textbf{0.929}}&\textcolor{tblue}{\textbf{0.744}}&\textcolor{tblue}{\textbf{0.519}}&\textcolor{tblue}{\textbf{0.330}}\\
% MODEL2 &  0.921 & 0.797 & 0.602 & 0.416 \\
\bottomrule
    \end{tabular}
    \end{adjustbox}
    \label{tab:emnlp}
\end{table}

\vspace{-4mm}

\subsubsection{Long-text generation}
% \vspace{-0.2cm}

%The EMNLP2017 WMT dataset is publicly available. 
To increase text generation difficulty, we use the whole sentences (with a maximum length of 50) without discarding low-frequency words on the EMNLP2017 WMT dataset. The experimental settings are the same as those used in the COCO dataset. Results are presented in Table~\ref{tab:emnlp}. Our method outperforms other baselines.

%   #################################################################

\vspace{-2mm}
\section{Analysis and Discussion}
% \vspace{-2mm}

To further evaluate the effectiveness and usefulness of our method, we carry out the following analyses and discussions on the more challenging dataset -- EMNLP WMT dataset, which has a longer sentence length, contains much more sentence structures and syntax than the COCO dataset. 

This section first conducts a human evaluation on Amazon Mechanical Turk, grammar check evaluation, and memorization evaluation. Then we carry out ablation studies to explore the effect of different training data sizes and the necessity of imitation learning. At last, we design two task-specific reference metrics to control vocabulary usage and nontrivial repetitions.  

% \vspace{-0.2cm}
\subsection{Human evaluations}

To further evaluate the generation quality, we conduct a human evaluation on Amazon Mechanical Turk; 3 judges are asked to rate over 100 randomly sampled sentences from each model with a scale from 1 to 5. The results shown in Table \ref{tab:human} indicate the outstanding performance of our proposed methods. 
% Please refer to Appendix \ref{app:results} for more details.

\vspace{-3mm}
\begin{table}[h!]
\centering
	\caption{{\small Human evaluation results.}}\label{tab:human}
	\vspace{-2mm}
	\begin{adjustbox}{scale=0.7}
	\begin{tabular}{l| c c c c}
	\midrule
	{Methods} & {{\em LeakRNN}} & {\em LeakRNN+RL} & {\em LeakRNN+SDA}& {Real} \\
	\midrule
	{Human scores $\uparrow$} & 2.97  &2.67& \textcolor{tblue}{\textbf{3.11}}  & 4.10 \\
    \bottomrule
    \end{tabular}
\end{adjustbox}
\end{table}

\begin{table*}
\parbox{.3\textwidth}{
\centering
\caption{{\small Grammar evaluation results.}}\label{tab:software evaluation}
\vspace{-2mm}
\begin{adjustbox}{scale=0.7}
\begin{tabular}{l|cc}
      \toprule
		Model   & Ave Len $\uparrow$   & CIR $\downarrow$    \\  
		\midrule
        Training Dataset & 27.526 & 2.583 \\
        {\em LeakRNN} & 23.990  & 9.880            \\  
        {\em LeakRNN+RL}  & 26.280  &  12.929 \\
        {\em LeakRNN+SDA}   & \textcolor{tblue}{\bf{27.850}} & \textcolor{tblue}{{\bf 9.874}}  \\
      \bottomrule
    \end{tabular}
    \end{adjustbox}
}
\hfill
\parbox{.31\textwidth}{
\centering
\caption{{\small Vocabulary usage control.}}\label{tab:vocabulary}
\vspace{-2mm}
    \begin{adjustbox}{scale=0.7}
    \begin{tabular}{l|cc}
		 \toprule
         & {BLEU2} $\uparrow$ & Occurrence  \\
         \midrule
        {\em LeakRNN}        & 0.914 &  0.2\% \\
		{\em LeakRNN+UN}   & 0.915 &  0.9\%\\
		{\em LeakRNN+SDA}      & \textcolor{tblue}{\bf{0.928}} &  0.5\% \\
    \bottomrule
	\end{tabular}
	\end{adjustbox}
}
\hfill
\parbox{.35\textwidth}{
\centering
\caption{{\small Repetition control.}}\label{tab:repeat}
    \vspace{-2mm}
    \begin{adjustbox}{scale=0.7}
    \begin{tabular}{l|cc}
		 \toprule
         & {BLEU3} $\uparrow$ & {self-BLEU3} $\downarrow$  \\
         \midrule
        {\em LeakRNN}        & 0.914 &  0.838\\
		{\em LeakRNN+UN}   & 0.912 &  0.800\\
		{\em LeakRNN+SDA}      & \textcolor{tblue}{\bf{0.920}} &  \textcolor{tblue}{\bf{0.816}} \\
    \bottomrule
	\end{tabular}
	\end{adjustbox}
}
\end{table*}

\vspace{-0.4cm}
\subsection{Grammar check evaluations}

We employ the grammar check tool~\footnote{{\small https://www.grammarly.com/}} for Grammar Check Evaluation, which is a professional software widely adopted for English writing checking in terms of {\em contextual spelling, grammar, punctuation, sentence structure, style, vocabulary enhancement}, measured by the {\em critical issue rate} (CIR). 
The results are shown in Table~\ref{tab:software evaluation}.
Please refer to Appendix~\ref{app:grammar} for more details.   

\vspace{-2mm}
\subsection{Memorization evaluations}

This experiment aims at comparing the generalization ability of different methods. To this end, we note that BigGAN \cite{brock2018large} carried out the nearest neighbor analysis on ImageNet \cite{imagenet_cvpr09} to shown their generator does not merely memorize data from the training set, which is a direct reflection of a model's generalization ability. As a result, we compare the BLEU score between our generated sentences and training dataset with {\em LeakRNN} and {\em LeakRNN+RL}. Figure~\ref{fig:memory} shows a better ability of our model for generating sentences instead of memorizing sentences directly from the training dataset.

\subsection{Improvement with different training data sizes}

% \vspace{-2mm}
\begin{table}[h]
\centering
\small
\caption{Performance comparison on different size of training dataset.}\label{tab:train_size}
    \vspace{-2mm}
    \begin{adjustbox}{scale=0.7}
    \begin{tabular}{l|ccc}
		 \toprule
         &Size& {BLEU2} $\uparrow$ & Increase \%  \\
         \midrule
        \multirow{2}{*}{5\%} & {\em LeakRNN}   & 0.8199 & \multirow{2}{*}{4.12\%} \\
		                     & {\em LeakRNN+SDA} & 0.8537 &\\
		 \midrule
		 \multirow{2}{*}{10\%} & {\em LeakRNN} & 0.8551 & \multirow{2}{*}{2.28\%} \\
		                     & {\em LeakRNN+SDA} & 0.8746 &\\
		 \midrule
		 \multirow{2}{*}{50\%} & {\em LeakRNN} & 0.9024 & \multirow{2}{*}{1.74\%} \\
		                     & {\em LeakRNN+SDA} & 0.9181 &\\
		  \midrule
		 \multirow{2}{*}{100\%} & {\em LeakRNN} & 0.9144 & \multirow{2}{*}{1.60\%} \\
		                     & {\em LeakRNN+SDA} & \textcolor{tblue}{\textbf{0.9290}} &\\
    \bottomrule
	\end{tabular}
	\end{adjustbox}
\end{table}

To test our method's performance with different sizes of the training set, we split the training dataset of EMNLP2017 to 5\%, 10\%, and 50\%, respectively. We compare our method with the MLE baseline {\em LeakRNN}. The results are shown in Table~\ref{tab:train_size}. Interestingly, with our data augmentation scheme, our method could boost performance over the model base more significantly on smaller training data. We speculate that since overfitting tends to be more severe when training on smaller datasets, our method could alleviate the overfitting problem seamlessly as a data augmented method.

\subsection{The necessity of imitation learning}

We conduct experiments to verify the necessity of imitation learning to mimic both the training data and self-augmented data. 
To this end, instead of leverage imitation learning to update the generator alternatively, we mix training data with 5\% the buffer $\mathcal{D}_B$ at each epoch and use these mixed data to update the generator after the {\em MLE Training stage}. 
Mixup \cite{zhang2017mixup} could be successfully applied to data augmentation for the image. We tried this method, but it did not gain better performance than EDA \cite{wei2019eda} in text augmentation. Then we did not include the result from this method in this article. 
Our results show that {\em LeakRNN} achieved 0.916 BLEU2 score, while the model updating generator used mixed data achieves 0.919, versus 0.943 with our {\em LeakRNN+SDA} model. The results indicate the benefits of using imitation learning.
% imitating both the training self-augmented data.

% For model without imitation learning, we will select top k generated sentences using the same metrics as our $MODEL2_2$ and add these selected sentences to our training dataset. We found that if we directly add Buffer $\mathcal{D}_B$ to training dataset, it will reduce the performance of our baseline model $MODEL2$. Therefore, we only add top 50\% sentences in $\mathcal{D}_B$. The model without imitation learning will achieve $0.919$ BLEU2 score.

\subsection{Vocabulary usage control}
It has been observed that generative models tend to generate common words too frequently, and rare words too infrequently, as compared with human distribution \cite{holtzman2018learning,welleck2019neural,li2019don,roller2020recipes}. 
We first analyze our training data distribution, which contains over 7 million words, with 5,722 unique characters in the dictionary. That means each word appears 1,260 times on average.
However, 20 words occur less than ten times, 43 words less than 20 times. 

% On the contrary, 8 words occurs more than 100,000 times. 
% The mean occurrences per word is 1260, the standard deviation is 72,963,235. It is very hard to find the rare words in the generated sentences. 

Our model could be used to control vocabulary usage during training due to the flexibility of defining a specific metric to select generated sentences. To this end, we collect characters with occurrences of less than 50 to be a special character list $\mathcal{C}_{\text{rare}}$. We define a new reference metric as: 
\begin{small}
\begin{align}\label{eq:reference vob usage}
 m(\sbb) \triangleq \text{BLEU}(\sbb) + 1/o^2,
\end{align}
\end{small}
where $o$ means occurrences times in the training dataset for words in the collection $\mathcal{C}_{\text{rare}}$. For example, the word "project" occurs 20 times in the training dataset. Once the generated sentences contain the word "project," we add 0.0025 to our reference metrics. If the generated sentence contains two special words, we add $1/o^2$ term twice.

As a comparison, we adopt the unlikelihood estimation \cite{welleck2019neural,li2019don,roller2020recipes} and combine it with the {\em LeakRNN} model (named {\em LeakRNN+UN}) to control vocabulary usage with a modified objective:
\begin{small}
\begin{equation*}
    J_g(\pi)= \sum_{t=1}^{T} \left\{ \begin{array}{ll}
         \log G_{\theta}\left(y_{t} | y_{1:t-1}\right) & \text{for } y_t \not\in  \mathcal{C}_{\text{rare}}\\
         \\
         \log G_{\theta}\left(y_{t} | y_{1:t-1}\right) + 1/o &  \text{for } y_t \in  \mathcal{C}_{\text{rare}}.
    \end{array}\right.
%\label{eq:unlike}
\end{equation*}
\end{small}

% \vspace{-0.3cm}
Table~\ref{tab:vocabulary} shows that our method could control the appearance of rare words to a certain extent, and can also improve the quality of generated sentences.
% \vspace{-0.5cm}

% \begin{table}[h!]
%   \centering
%     \caption{Vocabulary usage control.}\label{tab:vocabulary}
%     % \vspace{-2mm}
%     \begin{adjustbox}{scale=0.7}
%     \begin{tabular}[width=0.4\textwidth]{l|cc}
% 		 \toprule
%          & {BLEU2} $\uparrow$ & Occurrence rate  \\
%          \midrule
%         {\em LeakRNN}        & 0.914 &  0.2\% \\
% 		{\em LeakRNN+UN}   & 0.915 &  0.9\%\\
% 		{\em LeakRNN+SDA}      & 0.928 &  0.4\% \\
%     \bottomrule
% 	\end{tabular}
% 	\end{adjustbox}
% % 	\vspace{-2mm}
% \end{table}

% \vspace{-0.5cm}
\subsection{Nontrivial repetition control}

A related issue to vocabulary usage is that generative models also tend to repeat \cite{holtzman2019curious}. \cite{roller2020recipes} uses unlikelihood training to penalize 3-gram repetition in generated sentences. We use this model as a baseline, denoted as {\em LeakRNN+UN}. Again, 
we use equation~\eqref{eq:reference vob usage} as the reference metric, with $o$ representing the occurrence times of repeated 3-gram phases in a sentence. We adapt the self-BLEU3 score \cite{zhu2018texygen} to measure the repetition of 3-gram phases in the generated sentences.

% \begin{table}[h!]
%   \centering
%     \caption{Repetition control.}\label{tab:repeat}
%     % \vspace{-2mm}
%     \begin{adjustbox}{scale=0.85}
%     \begin{tabular}{l|cc}
% 		 \toprule
%          & {BLEU3} $\uparrow$ & {self-BLEU3} $\downarrow$  \\
%          \midrule
%         {\em LeakRNN}        & 0.914 &  0.838\\
% 		{\em LeakRNN+UN}   & 0.912 &  0.800\\
% 		{\em LeakRNN+SDA}      & 0.920 &  0.818 \\
%     \bottomrule
% 	\end{tabular}
% 	\end{adjustbox}
% % 	\vspace{-2mm}
% \end{table}

Table~\ref{tab:repeat} shows that unlikelihood training could significantly reduce the repetition rate at the cost of lowering generation quality, {\it i.e.}, it can help control repetition with the risk of saying something that makes less sense. On the contrary, our method could seamlessly control repetition while keeping generation quality.

% \vspace{-2mm}
\section{Conclusion}
% \vspace{-3mm}
We propose a general and flexible text augmentation method that could be easily incorporated into the popular MLE based methods to boosting model performance. Essentially, we propose to add a {\em Self Data Augmentation stage} after the standard {\em MLE Training stage}. At this stage, our generator alternatively imitates training sentences and self-augmented sentences selected by a reference metric. We show formally that doing so can enhance the generator to focus on higher-quality sentences via the proposed IRL formulation with self-enhancement. 
Moreover, by defining different reference metrics, we can emphasize some characteristics of the generated sentences, {\it e.g.}, vocabulary usage, and nontrivial repetition control.
Extensive experiments are conducted for both unconditional and conditional text-generation, verifying the effectiveness of our proposed data-augmentation framework for NLP.

%###############################################################################

\newpage
\newpage
% \bibliography{acl2020}
% \bibliographystyle{acl_natbib}
% \bibliographystyle{aaai21}
\bibliography{emnlp2020}

\begin{thebibliography}{44}
\providecommand{\natexlab}[1]{#1}
\providecommand{\url}[1]{\texttt{#1}}
\providecommand{\urlprefix}{URL }
\expandafter\ifx\csname urlstyle\endcsname\relax
  \providecommand{\doi}[1]{doi:\discretionary{}{}{}#1}\else
  \providecommand{\doi}{doi:\discretionary{}{}{}\begingroup
  \urlstyle{rm}\Url}\fi

\bibitem[{Ambrosio, Gigli, and Savar\'{e}(2005)}]{Ambrosio:book05}
Ambrosio, L.; Gigli, N.; and Savar\'{e}, G. 2005.
\newblock \emph{Gradient Flows in Metric Spaces and in the Space of Probability
  Measures}.
\newblock Lectures in Mathematics ETH Z{\"u}rich.

\bibitem[{Bahdanau et~al.(2017)Bahdanau, Brakel, Xu, Goyal, Lowe, Pineau,
  Courville, and Bengio}]{bahdanau2016actor}
Bahdanau, D.; Brakel, P.; Xu, K.; Goyal, A.; Lowe, R.; Pineau, J.; Courville,
  A.; and Bengio, Y. 2017.
\newblock An actor-critic algorithm for sequence prediction.
\newblock In \emph{ICLR}.

\bibitem[{Boyd et~al.(2011)Boyd, Parikh, Chu, Peleato, and
  Eckstein}]{Boyd:2011:DOS:2185815.2185816}
Boyd, S.; Parikh, N.; Chu, E.; Peleato, B.; and Eckstein, J. 2011.
\newblock Distributed Optimization and Statistical Learning via the Alternating
  Direction Method of Multipliers.
\newblock \emph{Found. Trends Mach. Learn.} 3(1): 1--122.
\newblock ISSN 1935-8237.
\newblock \doi{10.1561/2200000016}.
\newblock \urlprefix\url{http://dx.doi.org/10.1561/2200000016}.

\bibitem[{Brock, Donahue, and Simonyan(2018)}]{brock2018large}
Brock, A.; Donahue, J.; and Simonyan, K. 2018.
\newblock Large scale gan training for high fidelity natural image synthesis.
\newblock \emph{arXiv preprint arXiv:1809.11096} .

\bibitem[{Caccia et~al.(2018)Caccia, Caccia, Fedus, Larochelle, Pineau, and
  Charlin}]{caccia2018language}
Caccia, M.; Caccia, L.; Fedus, W.; Larochelle, H.; Pineau, J.; and Charlin, L.
  2018.
\newblock Language gans falling short.
\newblock \emph{arXiv preprint arXiv:1811.02549} .

\bibitem[{Deng et~al.(2009)Deng, Dong, Socher, Li, Li, and
  Fei-Fei}]{imagenet_cvpr09}
Deng, J.; Dong, W.; Socher, R.; Li, L.-J.; Li, K.; and Fei-Fei, L. 2009.
\newblock {ImageNet: A Large-Scale Hierarchical Image Database}.
\newblock In \emph{CVPR09}.

\bibitem[{Fadaee, Bisazza, and Monz(2017)}]{fadaee2017data}
Fadaee, M.; Bisazza, A.; and Monz, C. 2017.
\newblock Data augmentation for low-resource neural machine translation.
\newblock \emph{arXiv preprint arXiv:1705.00440} .

\bibitem[{Finn, Levine, and Abbeel(2016)}]{Finn:2016:GCL:3045390.3045397}
Finn, C.; Levine, S.; and Abbeel, P. 2016.
\newblock Guided Cost Learning: Deep Inverse Optimal Control via Policy
  Optimization.
\newblock In \emph{Proceedings of the 33rd International Conference on
  International Conference on Machine Learning - Volume 48}, ICML'16, 49--58.
  JMLR.org.
\newblock \urlprefix\url{http://dl.acm.org/citation.cfm?id=3045390.3045397}.

\bibitem[{Goodfellow et~al.(2014)Goodfellow, Pouget-Abadie, Mirza, Xu,
  Warde-Farley, Ozair, Courville, and Bengio}]{GoodfellowAMXWOCB:NIPS14}
Goodfellow, I.; Pouget-Abadie, J.; Mirza, M.; Xu, B.; Warde-Farley, D.; Ozair,
  S.; Courville, A.; and Bengio, Y. 2014.
\newblock Generative Adversarial Nets.
\newblock In \emph{Neural Information Processing Systems (NIPS)}.

\bibitem[{Guo et~al.(2018)Guo, Lu, Cai, Zhang, Yu, and Wang}]{GuoLCZYW:AAAI18}
Guo, J.; Lu, S.; Cai, H.; Zhang, W.; Yu, Y.; and Wang, J. 2018.
\newblock Long Text Generation via Adversarial Training with Leaked
  Information.
\newblock In \emph{AAAI}.

\bibitem[{Ho and Ermon(2016{\natexlab{a}})}]{NIPS2016_6391}
Ho, J.; and Ermon, S. 2016{\natexlab{a}}.
\newblock Generative Adversarial Imitation Learning.
\newblock In Lee, D.~D.; Sugiyama, M.; Luxburg, U.~V.; Guyon, I.; and Garnett,
  R., eds., \emph{Advances in Neural Information Processing Systems 29},
  4565--4573. Curran Associates, Inc.
\newblock
  \urlprefix\url{http://papers.nips.cc/paper/6391-generative-adversarial-imitation-learning.pdf}.

\bibitem[{Ho and Ermon(2016{\natexlab{b}})}]{ho2016generative}
Ho, J.; and Ermon, S. 2016{\natexlab{b}}.
\newblock Generative adversarial imitation learning.
\newblock In \emph{NIPS}.

\bibitem[{Holtzman et~al.(2019)Holtzman, Buys, Du, Forbes, and
  Choi}]{holtzman2019curious}
Holtzman, A.; Buys, J.; Du, L.; Forbes, M.; and Choi, Y. 2019.
\newblock The curious case of neural text degeneration.
\newblock \emph{arXiv preprint arXiv:1904.09751} .

\bibitem[{Holtzman et~al.(2018)Holtzman, Buys, Forbes, Bosselut, Golub, and
  Choi}]{holtzman2018learning}
Holtzman, A.; Buys, J.; Forbes, M.; Bosselut, A.; Golub, D.; and Choi, Y. 2018.
\newblock Learning to write with cooperative discriminators.
\newblock \emph{arXiv preprint arXiv:1805.06087} .

\bibitem[{Hu et~al.(2017)Hu, Yang, Liang, Salakhutdinov, and
  Xing}]{hu2017controllable}
Hu, Z.; Yang, Z.; Liang, X.; Salakhutdinov, R.; and Xing, E.~P. 2017.
\newblock Controllable text generation.
\newblock In \emph{ICML}.

\bibitem[{Kafle, Yousefhussien, and Kanan(2017)}]{kafle2017data}
Kafle, K.; Yousefhussien, M.; and Kanan, C. 2017.
\newblock Data augmentation for visual question answering.
\newblock In \emph{Proceedings of the 10th International Conference on Natural
  Language Generation}, 198--202.

\bibitem[{Kobayashi(2018)}]{kobayashi2018contextual}
Kobayashi, S. 2018.
\newblock Contextual augmentation: Data augmentation by words with paradigmatic
  relations.
\newblock \emph{arXiv preprint arXiv:1805.06201} .

\bibitem[{Li et~al.(2018)Li, Jia, He, and Liang}]{li2018delete}
Li, J.; Jia, R.; He, H.; and Liang, P. 2018.
\newblock Delete, retrieve, generate: A simple approach to sentiment and style
  transfer.
\newblock \emph{arXiv preprint arXiv:1804.06437} .

\bibitem[{Li et~al.(2019)Li, Roller, Kulikov, Welleck, Boureau, Cho, and
  Weston}]{li2019don}
Li, M.; Roller, S.; Kulikov, I.; Welleck, S.; Boureau, Y.-L.; Cho, K.; and
  Weston, J. 2019.
\newblock Don't Say That! Making Inconsistent Dialogue Unlikely with
  Unlikelihood Training.
\newblock \emph{arXiv preprint arXiv:1911.03860} .

\bibitem[{Liu et~al.(2018)Liu, Li, Tang, and Zhou}]{LiuLTZ:NIPS18}
Liu, Q.; Li, L.; Tang, Z.; and Zhou, D. 2018.
\newblock Breaking the Curse of Horizon: Infinite-Horizon Off-policy
  Estimation.
\newblock In \emph{NIPS}.

\bibitem[{Mikolov et~al.(2010)Mikolov, Karafi{\'a}t, Burget,
  {\v{C}}ernock{\`y}, and Khudanpur}]{mikolov2010recurrent}
Mikolov, T.; Karafi{\'a}t, M.; Burget, L.; {\v{C}}ernock{\`y}, J.; and
  Khudanpur, S. 2010.
\newblock Recurrent neural network based language model.
\newblock In \emph{Eleventh Annual Conference of the International Speech
  Communication Association}.

\bibitem[{Ng, Harada, and Russell(1999)}]{Ng:1999:PIU:645528.657613}
Ng, A.~Y.; Harada, D.; and Russell, S.~J. 1999.
\newblock Policy Invariance Under Reward Transformations: Theory and
  Application to Reward Shaping.
\newblock In \emph{Proceedings of the Sixteenth International Conference on
  Machine Learning}, ICML '99, 278--287. San Francisco, CA, USA: Morgan
  Kaufmann Publishers Inc.
\newblock ISBN 1-55860-612-2.
\newblock \urlprefix\url{http://dl.acm.org/citation.cfm?id=645528.657613}.

\bibitem[{Ng and Russell(2000)}]{Ng:2000:AIR:645529.657801}
Ng, A.~Y.; and Russell, S.~J. 2000.
\newblock Algorithms for Inverse Reinforcement Learning.
\newblock In \emph{Proceedings of the Seventeenth International Conference on
  Machine Learning}, ICML '00, 663--670. San Francisco, CA, USA: Morgan
  Kaufmann Publishers Inc.
\newblock ISBN 1-55860-707-2.
\newblock \urlprefix\url{http://dl.acm.org/citation.cfm?id=645529.657801}.

\bibitem[{Oh et~al.(2018)Oh, Guo, Singh, and Lee}]{oh2018self}
Oh, J.; Guo, Y.; Singh, S.; and Lee, H. 2018.
\newblock Self-imitation learning.
\newblock \emph{arXiv preprint arXiv:1806.05635} .

\bibitem[{Papineni et~al.(2002)Papineni, Roukos, Ward, and
  Zhu}]{papineni2002bleu}
Papineni, K.; Roukos, S.; Ward, T.; and Zhu, W.-J. 2002.
\newblock BLEU: a method for automatic evaluation of machine translation.
\newblock In \emph{Proceedings of the 40th annual meeting on association for
  computational linguistics}, 311--318. Association for Computational
  Linguistics.

\bibitem[{Ranzato et~al.(2016)Ranzato, Chopra, Auli, and
  Zaremba}]{ranzato2015sequence}
Ranzato, M.; Chopra, S.; Auli, M.; and Zaremba, W. 2016.
\newblock Sequence level training with recurrent neural networks.
\newblock In \emph{ICLR}.

\bibitem[{Rennie et~al.(2016)Rennie, Marcheret, Mroueh, Ross, and
  Goel}]{rennie2016self}
Rennie, S.~J.; Marcheret, E.; Mroueh, Y.; Ross, J.; and Goel, V. 2016.
\newblock Self-critical sequence training for image captioning.
\newblock In \emph{CVPR}.

\bibitem[{Roller et~al.(2020)Roller, Dinan, Goyal, Ju, Williamson, Liu, Xu,
  Ott, Shuster, Smith et~al.}]{roller2020recipes}
Roller, S.; Dinan, E.; Goyal, N.; Ju, D.; Williamson, M.; Liu, Y.; Xu, J.; Ott,
  M.; Shuster, K.; Smith, E.~M.; et~al. 2020.
\newblock Recipes for building an open-domain chatbot.
\newblock \emph{arXiv preprint arXiv:2004.13637} .

\bibitem[{Sennrich, Haddow, and Birch(2015)}]{sennrich2015improving}
Sennrich, R.; Haddow, B.; and Birch, A. 2015.
\newblock Improving neural machine translation models with monolingual data.
\newblock \emph{arXiv preprint arXiv:1511.06709} .

\bibitem[{Shen et~al.(2017)Shen, Lei, Barzilay, and Jaakkola}]{shen2017style}
Shen, T.; Lei, T.; Barzilay, R.; and Jaakkola, T. 2017.
\newblock Style transfer from non-parallel text by cross-alignment.
\newblock In \emph{NIPS}.

\bibitem[{Song et~al.(2018)Song, Ren, Sadigh, and Ermon}]{SongRSE:NIPS18}
Song, J.; Ren, H.; Sadigh, D.; and Ermon, S. 2018.
\newblock Multi-Agent Generative Adversarial Imitation Learning.
\newblock In \emph{NIPS}.

\bibitem[{Sudhakar, Upadhyay, and Maheswaran(2019)}]{sudhakar2019transforming}
Sudhakar, A.; Upadhyay, B.; and Maheswaran, A. 2019.
\newblock Transforming delete, retrieve, generate approach for controlled text
  style transfer.
\newblock \emph{arXiv preprint arXiv:1908.09368} .

\bibitem[{Sugiyama and Yoshinaga(2019)}]{sugiyama2019data}
Sugiyama, A.; and Yoshinaga, N. 2019.
\newblock Data augmentation using back-translation for context-aware neural
  machine translation.
\newblock In \emph{Proceedings of the Fourth Workshop on Discourse in Machine
  Translation (DiscoMT 2019)}, 35--44.

\bibitem[{Wang and Yang(2015)}]{wang2015s}
Wang, W.~Y.; and Yang, D. 2015.
\newblock That’s so annoying!!!: A lexical and frame-semantic embedding based
  data augmentation approach to automatic categorization of annoying behaviors
  using\# petpeeve tweets.
\newblock In \emph{Proceedings of the 2015 Conference on Empirical Methods in
  Natural Language Processing}, 2557--2563.

\bibitem[{Wang et~al.(2018)Wang, Chen, Wang, and Wang}]{wang2018no}
Wang, X.; Chen, W.; Wang, Y.-F.; and Wang, W.~Y. 2018.
\newblock No metrics are perfect: Adversarial reward learning for visual
  storytelling.
\newblock \emph{arXiv preprint arXiv:1804.09160} .

\bibitem[{Wei and Zou(2019)}]{wei2019eda}
Wei, J.~W.; and Zou, K. 2019.
\newblock Eda: Easy data augmentation techniques for boosting performance on
  text classification tasks.
\newblock \emph{arXiv preprint arXiv:1901.11196} .

\bibitem[{Welleck et~al.(2019)Welleck, Kulikov, Roller, Dinan, Cho, and
  Weston}]{welleck2019neural}
Welleck, S.; Kulikov, I.; Roller, S.; Dinan, E.; Cho, K.; and Weston, J. 2019.
\newblock Neural text generation with unlikelihood training.
\newblock \emph{arXiv preprint arXiv:1908.04319} .

\bibitem[{Williams(1992)}]{williams1992simple}
Williams, R.~J. 1992.
\newblock Simple statistical gradient-following algorithms for connectionist
  reinforcement learning.
\newblock \emph{Machine learning} 8(3-4): 229--256.

\bibitem[{Yang et~al.(2018)Yang, Hu, Dyer, Xing, and
  Berg-Kirkpatrick}]{yang2018unsupervised}
Yang, Z.; Hu, Z.; Dyer, C.; Xing, E.~P.; and Berg-Kirkpatrick, T. 2018.
\newblock Unsupervised text style transfer using language models as
  discriminators.
\newblock In \emph{Advances in Neural Information Processing Systems},
  7287--7298.

\bibitem[{Yu et~al.(2017)Yu, Zhang, Wang, and Yu}]{YuZWY:AAAI17}
Yu, L.; Zhang, W.; Wang, J.; and Yu, Y. 2017.
\newblock Seq{GAN}: Sequence Generative Adversarial Nets with Policy Gradient.
\newblock In \emph{AAAI}.

\bibitem[{Zhang et~al.(2017)Zhang, Cisse, Dauphin, and
  Lopez-Paz}]{zhang2017mixup}
Zhang, H.; Cisse, M.; Dauphin, Y.~N.; and Lopez-Paz, D. 2017.
\newblock mixup: Beyond empirical risk minimization.
\newblock \emph{arXiv preprint arXiv:1710.09412} .

\bibitem[{Zhang, Zhao, and LeCun(2015)}]{zhang2015character}
Zhang, X.; Zhao, J.; and LeCun, Y. 2015.
\newblock Character-level convolutional networks for text classification.
\newblock In \emph{Advances in neural information processing systems},
  649--657.

\bibitem[{Zheng et~al.(2019)Zheng, Zhou, Huang, Li, Dai, and
  Chen}]{zheng2019mirror}
Zheng, Z.; Zhou, H.; Huang, S.; Li, L.; Dai, X.-Y.; and Chen, J. 2019.
\newblock Mirror-Generative Neural Machine Translation.
\newblock In \emph{International Conference on Learning Representations}.

\bibitem[{Zhu et~al.(2018)Zhu, Lu, Zheng, Guo, Zhang, Wang, and
  Yu}]{zhu2018texygen}
Zhu, Y.; Lu, S.; Zheng, L.; Guo, J.; Zhang, W.; Wang, J.; and Yu, Y. 2018.
\newblock Texygen: A Benchmarking Platform for Text Generation Models.
\newblock \emph{arXiv preprint arXiv:1802.01886} .

\end{thebibliography}
% \bibliography{emnlp2020}

\newpage
\clearpage
\appendix

\section{Preliminaries about Inverse Reinforcement Learning}
\label{app:preliminaries IRL}
% \subsection{Inverse reinforcement learning}
% \vspace{-0.1cm}
% An alternative way to write the RL objective equation~\ref{eq:rl} is

% J(\pi) = \sum_{t=1}^{T} \mathbb{E}_{y_{t} \sim G_{\theta}\left(y_{t} | y_{1: t-1}\right)}[Q_{D_{\phi}}(y_{1: t-1}, y_{t})]

An alternative way \cite{ho2016generative,LiuLTZ:NIPS18} to write the RL objective (equation~\ref{eq:rl}) is: 
% \vspace{-0.2cm}
\begin{small}
\begin{align}\label{eq:dpi}
J(\pi) & = \mathbb{E}_{\sbb_t\sim d_{\pi}(\cdot), y_{t} \sim G_{\theta}(y_{t} | y_{1: t-1})}\left[Q_{D_{\phi}}^{G_{\theta}}(y_{1: t-1}, y_{t})\right], ~~\text{with}~~\nonumber \\
   d_{\pi}(\sbb) &\triangleq \lim_{T \rightarrow \infty}(\sum_{t=0}^{T}\gamma^{t-1}d_{\pi, t}(\sbb)) / (\sum_{t=0}^T \gamma^t),
\end{align} \par \vspace{-0.3cm}
\end{small}
where $d_{\pi, t}(\sbb)$ is the distribution of $\sbb_t$ when executing policy $\pi$ starting from some initial state distribution; and 
$d_{\pi}(\sbb)$ is the state marginal distribution.

In order to unify the symbols in our formula with the symbols in Reinforcement Learning, we then use $r(\sbb_t, \ab_t)$ to represent $Q_{D_{\phi}}^{G_{\theta}}(y_{1: t-1}, y_{t})$, $\ab_t\sim \pi(\cdot|\sbb_t)$ to represent $y_{t} \sim G_{\theta}(y_{t} | y_{1: t-1})$ in the following theories.

The {\em occupancy measure} $\rho(\sbb, \ab)$ is defined as the stationary joint distribution of $(\sbb, \ab)$ under policy $\pi$, {\it i.e.}, $\rho(\sbb, \ab) \triangleq d_{\pi}(\sbb)\pi(\ab|\sbb)$. Thus, given an environment, there is a one-to-one correspondence between the policy $\pi$ and occupancy measure $\rho$ \cite{ho2016generative}. Consequently, RL can be equivalently rewritten as
\begin{align}~\label{eq:rl_occupancy}
     \pi^* \triangleq \RL(\pi) = 
	 \arg\max_{\pi \in \Pi }
	 \mathbb{E}_{(\sbb, \ab) \sim \rho}\left[r(\sbb, \ab)\right]~.
\end{align}

To formulate text generation as RL, we may consider the process as an agent taking actions to generate next words, conditioned on the states of observed words, {\it i.e.},
\begin{align}
     \sbb_t = w_{<t}~~\text{and}~~\ab_t = w_t~.
\end{align}
% the state $\sbb_t$ corresponds to the (sub-)sentence $\sbb_t \triangleq w_{<t}$, and the action is the next generated word $\ab_t \triangleq w_t$. 
%
% Different from the traditional RL, the transition probability $P_s$ is deterministic given the current state-action pair in text generation. The policy generates the next word given the previous generated words; and the reward measure how good the next word is from the policy. 
%
Hence, the performance of RL-based text generation largely depends on the design of the reward $r(\sbb, \ab)$ in~\eqref{eq:rl_occupancy}. Previous methods directly consider the rewards as the hand-crafted metrics between ground-truth sentences and generated sentences, such as BLEU~\citep{ranzato2015sequence}, and improve its learning efficiency with variance reduction techniques~\cite{bahdanau2016actor, rennie2016self}. 
However, it is well-acknowledged that text evaluation is an extremely challenging problem, and the current metrics are imperfect for various reasons~\citep{wang2018no}. A direct use of these metrics in RL may lead to sub-optimal text generators.

Inverse reinforcement learning (IRL), by contrast, aims to learn the optimal reward function from the {\em expert trajectories} $\mathcal{D}_E$, which is represented as a training dataset in text generation. Typically, $\mathcal{D}_E$ consists of state-action pairs generated from an expert policy $\pi_E$, with the corresponding occupancy measure $\rho_E$. The goal of IRL is to recover the optimal reward function $r^*(\cdot, \cdot)$ as well as the corresponding policy $\pi^*$. %For convenience, we further define a cost function $c$ as the negative reward function, {\it i.e.}, $c(\sbb, \ab) \triangleq -r(\sbb, \ab)$. 
% In the following, we will describe IRL and our method in the case of discrete state and action spaces for simplicity, although they can be extended to the continuous-space case. 
Formally, we have for IRL\footnote{The original goal of IRL is to find the optimal reward function. In this paper, we generalize the definition to finding both optimal reward and policy.}.

\vspace{-0.5cm}
{\small\begin{align*}
	&\{r^*, \pi^*\} \triangleq \IRL(\pi_E)\\ 
	&= 
	 \arg\max_{r\in\mathbb{R}^{\mathcal{S}\times \mathcal{A}}} \sum_{\sbb, \ab}\rho_{E}(\sbb, \ab)r(\sbb, \ab)\\ 
	 & -[\max_{\pi\in\Pi}H(\pi) + \sum_{\sbb, \ab}\rho(\sbb, \ab)r(\sbb, \ab)] \\
	& = \arg\max_{r\in\mathbb{R}^{\mathcal{S}\times \mathcal{A}}} \min_{\pi\in\Pi} ~~\mathcal{L}(\pi, r)\\
	& \triangleq \! -H(\pi)\! +\! \sum_{\sbb, \ab}\rho_{E}(\sbb, \ab)r(\sbb, \ab)
	\!-\! \sum_{\sbb, \ab}\rho(\sbb, \ab)r(\sbb, \ab).
\end{align*}}
Intuitively, the objective enforces the expert policy $\pi_E$ to induce higher rewards (the ``$\max$'' part) than all other policies (the ``$\min$'' part). Note this setting is sub-optimal if the given data is noisy, as the learned policy is expected to perform worse than the expert policy (and equal to the expert policy at optimality). Besides, the IRL objective is ill-defined \cite{Ng:2000:AIR:645529.657801,Finn:2016:GCL:3045390.3045397}, which often induces multiple valid solutions to the problem because the solution space is too flexible. For example, one can assign an arbitrary reward to trajectories not visited by the expert, as long as these trajectories yield lower rewards than the expert trajectories \cite{SongRSE:NIPS18}. To address these problems, some constraints are typically placed on the reward functions, {\it e.g.}, a convex reward functional, $\psi: \mathbb{R}^{\mathcal{S}\times\mathcal{A}} \rightarrow \mathbb{R}$, is usually introduced as a regularizer \cite{ho2016generative}. Finally, the IRL is formulated as
\begin{align}\label{eq:IRL}
\{\pi^*, r^*\} =& \arg \min_{\pi\in\Pi}\max_{r\in\mathbb{R}^{\mathcal{S}\times \mathcal{A}}} 
\mathcal{L}(\pi, r)  - \psi(r).
\end{align}\par 
\vspace{-0.2cm}

\section{Self-Improved Inverse RL}
\label{app:theory behind our method}
In this paper, we advocate that IRL is a more natural framework to formulate text generation. Each training sentence sample plays the role of an expert trajectory.  
The goal becomes recovering the optimal reward function $r$, as well as the corresponding language model (the optimal policy $\pi$).

%
% The traditional regularization in IRL only considers constraining the model capacity of the policy, such as linear, convex, and adversarial regularizers~\cite{ho2016generative}, which makes it hard to be adapted to diverse tasks. 
Natural languages own intrinsic characteristic based on syntax and semantics. 
To leverage this prior knowledge, we propose to incorporate user-defined evaluation measures into $\psi(r)$ so that the learned policy aligns well with the pre-defined metrics when generating unseen sentences. 
Specifically, we define $\psi(r)$ in \eqref{eq:IRL} to be a mixture of a {\it capacity regularizer}  $\psi_1(r)$ and a {\it self-guided regularizer}  $\psi_0(r)$:
\begin{align}\label{eq_mix_regularier}
	\psi(r) = -\lambda \psi_0(r) + \psi_1(r)~, % 
\end{align}
where $\lambda > 0$ is the weighting hyper-parameter to balance two terms. 

\vspace{-2mm}
\paragraph{Capacity Regularizer}~
It has been shown in~\cite{ho2016generative} that the adversarial regularizer can provide better generalization.  
To imitate the expert policy, we adopt the {\it capacity regularizer}  $\psi_1(r)$ in GAIL~\cite{ho2016generative}. 

% \cy{Write out the form in the appendix}
% , which defines $\psi_1$ with the following form:
% \begin{align}\label{eq_capacity_regularizer}
% 	\psi_1(r) \triangleq \left\{
% 		\begin{array}{ll}
% 			\mathbb{E}_{\pi_E}\left[g(r(\sbb, \ab))\right] & \mbox{if } r(\sbb, \ab) \geq 0\\
% 			+\infty & \mbox{otherwise}
% 		\end{array}~,
% 	\right.
% \end{align}

\paragraph{Self-Guided Regularizer}~To promote generating suboptimal demonstrations that are well aligned with the diverse tasks, we propose the {\it self-guided regularizer} as a weighted linear function over rewards: 

\vspace{-0.3cm}
\begin{align}\label{eq_semantic_regularier}
	 \psi_0(r) = \sum_{\sbb, \ab}q(\sbb, \ab)r(\sbb, \ab) % ~~\text{with}~~
\end{align}\par \vspace{-0.3cm}
\noindent where $q(\sbb, \ab)$ is a probability function constructed via evaluation metrics (to be defined below). Generally, this regularizer encourages higher rewards in regions where $q(\sbb, \ab)$ are large.
In text generation, the evaluation metrics are used to evaluate certain quality aspects of a generated sentence, such as the BLEU or ROUGE scores. Formally, we define the notation of {\em reference metric}.

\begin{definition}[Reference Metric]
    A reference metric is defined as a function,  $m:\mathcal{S}\rightarrow \mathbb{R}$, mapping a sentence $\sbb \in \mathcal{S}$ to a real value measuring the goodness of the sentence. 
\end{definition}

For simplicity, we assume that a higher value of $m(\sbb)$ indicates higher quality of the sentence $\sbb$.
Note that the standard IRL with a fixed pool of expert trajectories does not necessarily guarantee to learn a reward function that matches the reference metric. 
% In another word, a high-rewarded sentence does not necessarily guarantee a high value measured by the reference metrics. 
We conjecture it may be caused by the fact that the expert trajectories alone can be noisy and unrepresentative. 

To {\em enhance} the state space in terms of the reference metric, we focus on a sub-space, $\bar{\mathcal{S}} \subseteq \mathcal{S}$, with high reference-metric values, {\it i.e.}, $\bar{\mathcal{S}} \triangleq \{\sbb \in \mathcal{S}: m(\sbb) > m^*\}$ for some threshold $m^*$. Next, the concentrated distribution $q(\sbb, \ab)$ is defined as

\vspace{-0.2cm}
{\small
\begin{align*}%\label{eq:dpi_1}
\hspace{-0.3cm}q(\sbb, \ab)& =
	 \frac{1}{Z}d_{\pi}(\sbb)\pi(\ab|\sbb)\delta(m(\sbb) > m^*)  \triangleq \bar{d}_{\pi}(\sbb)\pi(\ab|\sbb)~,
\end{align*}\par \vspace{-0.2cm}}
\noindent where the indication function $\delta(m(\sbb) > m^*) = 1$ if $m(\sbb) > m^*$ and 0 otherwise; $Z$ is the normalizer to ensure $q$ to be a proper distribution. In contrast to the original occupancy measure, $\rho(\sbb, \ab) = d_{\pi}(\sbb)\pi(\ab|\sbb)$ for all state-action pair in $\sbb \in \mathcal{S}$, $q(\sbb, \ab)$ is more concentrated towards regions with higher reference-metric values.
In another word, $q(\sbb, \ab)$ is larger than $\rho(\sbb, \ab)$ as the probabilities are only defined on the sub-space $\bar{\mathcal{S}}$ with high reference-metric values.
This method is related to reward shaping \cite{Ng:1999:PIU:645528.657613}, however, in a more principled manner. Unlike traditional reward-shaping strategy where the policy is directly trained to maximize certain heuristic thus it could be noisy, the uncertainty in $q(\sbb, \ab)$ still allows an automatic and softened exploration and exploitation trade-off.

% Our goal now becomes to integrate the enhanced probability distribution $q$ into the IRL framework. We find it convenient to directly incorporate $q$ into the regularized functional $\psi$ in IRL. This can be achieved as the linear functional emphasizes the reward function to focus on $\bar{\mathcal{S}}$.

By plugging~\eqref{eq_mix_regularier} and ~\eqref{eq_semantic_regularier} into~\eqref{eq:IRL}, we formulate our objective as an optimization problem to find the optimal reward $\r^*$ and the optimal policy $\pi^*$:

\vspace{-0.0cm}
{\small
\begin{align}\label{eq:SEIRL}
	&\{\pi^*, r^*\} = \arg\min_{\pi\in\Pi}\max_{r\in\mathbb{R}^{\mathcal{S}\times \mathcal{A}}} -2H(\pi)- \psi_1(r)  +   \\
& \!\! \sum_{\sbb, \ab}q(\sbb, \ab)r(\sbb, \ab) \!\! + \!\! \sum_{\sbb, \ab}\rho_E(\sbb, \ab)r(\sbb, \ab) \!\! -  \!\! 2\sum_{\sbb, \ab}\rho(\sbb, \ab)r(\sbb, \ab),\nonumber
\end{align}}
\noindent where we have set $\lambda = 1$ in~\eqref{eq_mix_regularier}, and the factor ``2'' is introduced for formulation convenience. 
% To emphasize the distinction of enhancing the expert trajectories under the guidance of reference metrics, we call our method {\it Self-Improved Inverse Reinforcement Learning} (SE-IRL).  

\subsection{A Two-step Algorithm for the Self Data Augmentation stage}
Due to the specific definition of $q(\sbb, \ab)$, solving \eqref{eq:SEIRL} directly is difficult. %\footnote{Though one can apply SGD for direct optimization, the performance may not be good due to the complexity of the problem.}. 
We reformulate \eqref{eq:SEIRL} with an augmented policy $\pi_1$:

\vspace{-0.0cm}
{\small\begin{align}\label{eq:SEIRL1}
	% \tilde{\mathcal{L}}= 
	 & \min_{\pi_1, \pi\in\Pi}\max_{r\in\mathbb{R}^{\mathcal{S}\times \mathcal{A}}} -H (\pi_1) + \sum_{\sbb, \ab}(q(\sbb, \ab) - \rho(\sbb, \ab))r(\sbb, \ab)\nonumber \\
	& - H(\pi) - \psi_1(r) + \sum_{\sbb, \ab}(\rho_E(\sbb, \ab) - \rho(\sbb, \ab))r(\sbb, \ab),
	\nonumber\\
	&~~~~~~~~~~~~~~~~~~~~\mbox{s.t. }~~ r_1 = r,~~ \pi_1 = \pi
\end{align}}

% \vspace{-0.5cm}
Objective \eqref{eq:SEIRL1} now allows us to decompose the original problem into two sub-problems. Adopting ideas from ADMM \cite{Boyd:2011:DOS:2185815.2185816}, we decompose \eqref{eq:SEIRL1} into two sub-problems, an {\it imitating self augmented data step} and an {\it imitating training data step}:

% \vspace{-0.2cm}
{\small\begin{flalign}\label{eq:SEIRL-1}
	& \text{imitating self augmented data step: }\\ \nonumber &\min_{\pi_1\in\Pi}\max_{r\in\mathbb{R}^{\mathcal{S}\times \mathcal{A}}}\mathcal{L}_1(\pi_1, r_1) 
	\triangleq -H(\pi_1) \\
	&+ \sum_{\sbb, \ab}(q(\sbb, \ab) - \rho(\sbb, \ab))r(\sbb, \ab) + \lambda W_1(\pi, \pi_1) \nonumber
\end{flalign}}

\vspace{-0.7cm}
{\small\begin{flalign}\label{eq:SEIRL-2}
    & \text{imitating training data step: } \\ \nonumber &\min_{\pi\in\Pi}\max_{r\in\mathbb{R}^{\mathcal{S}\times \mathcal{A}}}\mathcal{L}_2(\pi, r) \triangleq - H(\pi)\\ 
    &- \psi_1(r)+ \sum_{\sbb, \ab}(\rho_E(\sbb, \ab) - \rho(\sbb, \ab))r(\sbb, \ab) 
    + \lambda W_1(\pi, \pi_1)~. \nonumber
\end{flalign}}

\vspace{-0.1cm}
The regularizer $W_1(\pi, \pi_1)$ above is introduced to accommodate the constraint of $\pi_1 = \pi$ in the original problem \eqref{eq:SEIRL1}. 
% Here $\lambda_1$ is a weighting hyper-parameters\footnote{The parameter, in principle, should also be optimized. However, we fix it in practice for simplicity.}. 
The two phases \eqref{eq:SEIRL-1} and \eqref{eq:SEIRL-2} are solved alternatively. By parametrizing all the $\pi$, $\pi_1$ and $r$ with DNNs, they can be updated by solving the min-max problems similar to GAIL \cite{NIPS2016_6391}. 
% We defer the detailed derivations to Appendix \ref{app:IL}, as these are standard techniques. 
The only difference lies on the $W_1$ regularizer, where the gradients of policy parameters need to be derived. Fortunately, this can be solved by the following theorem.

\begin{theorem}\label{theo:gradient}
    Let $\pi$ be parameterized with $\thetab$, denoted as $\pi_{\thetab}$. The gradient of $W_1(\pi_{\thetab}, \pi_1)$ w.r.t.\! $\thetab$ can be calculated as
    {\small
    % \begin{align*}
    %     \nabla_{\theta}&W_1(\pi_\theta, \pi_1) = \int \frac{\delta W_1}{\delta \pi_{\theta}}(\pi_{\theta},\pi_1) \cdot \nabla_{\theta}\pi_{\theta}(Y)\mathrm{d}Y \\
    %     & = \mathbb{E}_{Y\sim \pi_{\theta}}\left[\frac{\delta W_1}{\delta \pi_{\theta}}(\pi_{\theta})(Y) \cdot\nabla_{\theta}\log\pi_{\theta}(Y)\right]~,
    % \end{align*}}
    % % to reviewer
    \begin{align*}
        \nabla_{\theta}&W_1(\pi_\theta, \pi_1) = \int \frac{\delta W_1}{\delta \pi_{\theta}}(\pi_{\theta},\pi_1) \cdot \nabla_{\theta}\pi_{\theta}(Y) \ dY \\
        &= \int \frac{\delta W_1}{\delta \pi_{\theta}}(\pi_{\theta},\pi_1) \cdot \frac{\nabla_{\theta}\pi_{\theta}(Y)}{\pi_{\theta}(Y)} \cdot \pi_{\theta}(Y) \ dY \\
        &= \int \frac{\delta W_1}{\delta \pi_{\theta}}(\pi_{\theta},\pi_1) \cdot \log\pi_{\theta}(Y) \cdot \pi_{\theta}(Y) \ dY \\
        & = \mathbb{E}_{Y\sim \pi_{\theta}}\left[\frac{\delta W_1}{\delta \pi_{\theta}}(\pi_{\theta})(Y) \cdot\nabla_{\theta}\log\pi_{\theta}(Y)\right]
    \end{align*}}
    %\vspace{-0.1cm}
    \noindent where the first equality follows by the chain rule, with $\frac{\delta W_1}{\delta \pi_{\theta}}(\pi_{\theta})$ defined as the first variation of $W_1$ \cite{Ambrosio:book05}. Furthermore, rewrite $W_1(\pi_{\thetab}, \pi_1)$ in its dual form as: $W_1(\pi_{\thetab}, \pi_1) = \max_{\|D\|_{\text{Lip}} < 1} \mathbb{E}_{Y \sim \pi_{\thetab}}\left[D(Y)\right] - \mathbb{E}_{Y^\prime \sim \pi_1}\left[D(Y^\prime)\right]$, with the optimal discriminator denoted as $D^*(\cdot)$. Then $\nabla_{\theta}W_1(\pi_{\theta}, \pi_1)$ can be calculated as
    {\small\begin{align}\label{eq:update_theta}
        & \nabla_{\theta}W_1(\pi_{\theta}, \pi_1) = \mathbb{E}_{Y\sim \pi_{\theta}}[D^*(Y)\nabla_{\theta}\log \pi_{\theta}(Y)] \\
        & + \mathbb{E}_{Y \sim \pi_{\theta}, Y^\prime \sim \pi_1}\left[\nabla_{\theta}Y\frac{\partial D^*(Y)}{\partial Y}-\nabla_{\theta}Y^\prime\frac{\partial D^*(Y^\prime)}{\partial Y^\prime}\right] \nonumber
    \end{align}}
\end{theorem}

\vspace{-0.1cm}
Theorem~\ref{theo:gradient} applies to the parameterized $\pi_1$ as well. It is now clear that to calculate gradients of the $W_1$ term, one $\RN{1})$ first draw samples from the two policies to update the discriminator $D$; $\RN{2})$ then use \eqref{eq:update_theta} to calculate $\nabla_{\theta}W_1$. This consequently allows us to optimize \eqref{eq:SEIRL-1} and \eqref{eq:SEIRL-2} by gradient descent. It is worth noting that when the two policies in the $W_1$ arguments are close enough to each other, the second part of the RHS of \eqref{eq:update_theta} will become relatively small, allowing us to drop it for computational efficiency. The whole algorithm is illustrated in Algorithm 1.

\section{Proof of Theorem~\ref{theo:gradient}}\label{app:theory 1}
% \begin{proof}
    The gradient of $W_1$ can be expanded, based on the definition of functional derivative, as:
    \begin{align*}
    \nabla_{\theta}&W_1(\pi_\theta, \pi_{1}) = \int \frac{\delta W_1}{\delta \pi_{\theta}}(\pi_{\theta},\pi_{1}) \cdot \nabla_{\theta}\pi_{\theta}(Y)\mathrm{d}Y\nonumber \\
    & = \mathbb{E}_{Y\sim \pi_{\theta}}\left[\frac{\delta W_1}{\delta \pi_{\theta}}(\pi_{\theta},\pi_{1}) \cdot\nabla_{\theta}\log\pi_{\theta}(Y)\right]~,
    \end{align*}
    where the first equality follows by the chain rule, with $\frac{\delta W_1}{\delta \pi_{\theta}}(\pi_{\theta})$ defined as the first variation of the functional $W_1$ \cite{Ambrosio:book05}.
    
    To prove the second claim, by substituting the optimal discriminator $D^*$ into the dual formula of $W_1$ and noting $D^*$ is implicitly dependent on $\theta$, we have
    {\small
    \begin{align*}
        &W_1(\pi_{\thetab}, \pi_1) = \mathbb{E}_{Y \sim \pi_{\thetab}}\left[D^*(Y)\right] - \mathbb{E}_{Y^\prime \sim \pi_1}\left[D^*(Y^\prime)\right]~.
    \end{align*}
    \begin{align*}
        & \Rightarrow \frac{\partial W_1(\pi_{\theta}, \pi_1)}{\partial \theta}\\
        & = \frac{\partial}{\partial \theta}\int D^*(Y)\pi_{\theta}(Y)\mathrm{d}Y - \mathbb{E}_{Y^\prime \sim \pi_1}\left[\nabla_{\theta}D^*(Y^\prime)\right] \\
        &= \int \nabla_{\theta}\pi_{\theta}(Y) D^*(Y)\mathrm{d}Y + \int \nabla_{\theta}D^*(Y) \pi_{\theta}(Y) - \\
        & ~~~~~~ \mathbb{E}_{Y^\prime \sim \pi_1}\left[\nabla_{\theta}D^*(Y^\prime)\right] \\
        &= \mathbb{E}_{Y\sim \pi_{\theta}}[D^*(Y)\nabla_{\theta}\log \pi_{\theta}(Y)] + \mathbb{E}_{Y \sim \pi_{\theta}}\left[\nabla_{\theta}D^*(Y)\right] - \\
        & ~~~~~ \mathbb{E}_{Y^\prime \sim \pi_1}\left[\nabla_{\theta}D^*(Y^\prime)\right] \\
        &= \mathbb{E}_{Y\sim \pi_{\theta}}[D^*(Y)\nabla_{\theta}\log \pi_{\theta}(Y)] + \\
        & ~~~~~ \mathbb{E}_{Y \sim \pi_{\theta}, Y^\prime \sim \pi_1}\left[\nabla_{\theta}Y\frac{\partial D^*(Y)}{\partial Y}-\nabla_{\theta}Y^\prime\frac{\partial D^*(Y^\prime)}{\partial Y^\prime}\right]~,
    \end{align*}}
    which completes the proof. 
    %where the last equality follows because the gradient vanishes at equilibrium (optimal solution).
% \end{proof}

\section{More Details of Imitation Learning}\label{app:IL}
% \subsection{Phase \rom{2}: Imitation Learning}\label{sec:IL}

This describes our solution to the {\it imitation learning} phase \eqref{eq:SEIRL-2}. According to \cite{ho2016generative}, let $D(\sbb, \ab) \triangleq \exp(-r(\sbb, \ab))$, then \eqref{eq:SEIRL-2} can be reformulated as
{\small\begin{align}\label{eq:SEIRL-2-1}
	&\hspace{-0.2cm}\{\pi, r\}=\arg\min_{\pi\in\Pi}\max_{r\in\mathbb{R}^{\mathcal{S}\times \mathcal{A}}} - H(\pi) \\
	&\hspace{-0.2cm} +\lambda W_1(\pi_1, \pi) + \mathbb{E}_{\pi}\left[\log D(\sbb, \ab)\right] + \mathbb{E}_{\pi_E}\left[\log(1 - D(\sbb, \ab))\right]  \nonumber
\end{align}}
As in the standard GAIL algorithm, \eqref{eq:SEIRL-2-1} can be solved by alternating between the following two optimization problems.

\paragraph{Reward update}
We first parameterize $r$ with $\phib$ implemented as a deep neural network (DNN), which takes a state-action pair as input and outputs a real value, {\it i.e.}, $r_{\phib}(\cdot, \cdot) \triangleq -\log D(\cdot, \cdot)$. Consequently, the reward network can be optimized as in the standard GAIL framework with the following objective:

\begin{align}\label{eq:op_phi}
	\phib^* = \arg \max_{\phib} \mathbb{E}_{\pi}\left[\log D_{\phib}(\sbb, \ab)\right] +\\ \mathbb{E}_{\pi_E}\left[\log(1 - D_{\phib}(\sbb, \ab))\right]
\end{align}\par

where the expectations over $\pi$ and $\pi_E$ are approximated by samples from the current policy $\pi$ (which is initialized with $\pi_1$ learned in the {\it self-improvement} phase) and the training data, respectively.

\paragraph{Policy update}
Given the learned reward network $r_{\phib}$, optimizing the policy can be done by standard policy-gradient-based methods, with the following objective:

\begin{align}\label{eq:SEIRL-2-3}
	\pi = \arg\max_{\pi\in\Pi} H(\pi) + \mathbb{E}_{\pi}\left[r_{\phib}(\sbb, \ab)\right] - \lambda W_1(\pi_1, \pi)~. 
\end{align}

\paragraph{Capacity Regularizer}~
It has been shown in~\citep{ho2016generative} that the adversarial regularizer can provide better generalization.  
To imitate the expert policy, we adopt the regularizer in GAIL~\cite{ho2016generative}, which defines $\psi_1$ with the following form:
\begin{align*}
	\psi_1(r) \triangleq \left\{
		\begin{array}{ll}
			\mathbb{E}_{\pi_E}\left[g(r(\sbb, \ab))\right] & \mbox{if } r(\sbb, \ab) \geq 0\\
			+\infty & \mbox{otherwise}
		\end{array}~,
	\right.
\end{align*}

\vspace{-4mm}
\begin{align*}
\text{where}\ g(x) = \begin{dcases*}-x-\log(1-e^x) & if $x < 0$ \\ +\infty & otherwise\end{dcases*}
\end{align*}

\section{Unsupervised Style Transfer}\label{app:style}
Our framework can also be extended to the task of style transfer. The key changes needed to make are: $\RN{1})$ Define the generator as a conditional distribution, instead of the unconditional distribution used for text generation; $\RN{2})$ Define the buffer $\mathcal{D}_B$ to store two bins of sentences, each corresponding to one style; $\RN{3})$ Include some additional reconstructed loss in order to preserve the reversibility of the learned policy. Detailed descriptions are given in Appendix \ref{app:style}.

We apply the method for non-parallel text-style-transfer, where the goal is to transfer one sentence in one style ({\it e.g.}, positive) to a similar sentence but with a different style ({\it e.g.}, negative). 
For a fair comparison, we use the same data and its split method as in \cite{shen2017style}. To measure whether the original test sentences have been transferred to the desired sentiment, we follow the settings of \cite{shen2017style} and employ a pretrained CNN classifier, which achieves an accuracy of $97.4\%$ on the validation set, to evaluate the transferred sentences. 
The results are presented in Table~\ref{tab:tranaccuracy}. Note that both \cite{li2018delete} and \cite{sudhakar2019transforming} are supervised approaches. It is observed that our method obtains much higher accuracy than other methods, including the controllable text generation method \cite{hu2017controllable}, while maintaining a comparable BLEU score. Some example-sentences are listed in Appendix \ref{app:results}.
% \vspace{-0.1cm}

\begin{table}[h!]
   \centering
    \caption{Style Transfer results on test set.}\label{tab:tranaccuracy}
    % \vspace{-0.1cm}
    \begin{adjustbox}{scale=0.7}
    \begin{tabular}{c|cc}
		 \toprule
         & {Accuracy} $\uparrow$ & {BLEU} $\uparrow$  \\
         \midrule
        VAE \cite{shen2017style} & 23.2\% & -\\
		Shen et al.~\cite{shen2017style} & 78.4\% & 7.8\\
		ST~\cite{yang2018unsupervised} & 83.3\%&56.2 \\
		
		Hu et al.~\cite{hu2017controllable} & 84.5\% & 58.7\\
		D\&R~\cite{li2018delete} & 89.3\% & 58.0 \\
        B-GST \cite{sudhakar2019transforming} & 87.3\% & 58.8 \\
        G-GST \cite{sudhakar2019transforming}& 78.3\% & 56.9 \\
		{\em SDA} & \bf{89.9\%} & 58.6 \\
    \bottomrule
	\end{tabular}
	\end{adjustbox}
% 	\vspace{-2mm}
\end{table}

Some example style-transfer sentences are listed in Table \ref{tab:styple transfer}.

\section{More Details of Experiments}\label{app:results}

\paragraph{Data statistics}
Table~\ref{tab:dataset} summarizes statistics of the data used in our experiments.

\begin{table}[!htb]
	%\vspace{-0.5cm}
	\centering
	\caption{Summary statistics for the datasets used in the unconditional text generation experiments}
	\begin{adjustbox}{scale=0.8}
	\begin{tabular}{c|c|c|c|c}
		\hline\toprule[1.2pt]
		Dataset            & Train   & Test   & $|\text{Voc}|$ & Avg Len \\  \midrule[0.7pt]
Synthetic 1        & 10,000  & 10,000 & 5,000      & 20     \\  
Synthetic 2        & 10,000  & 10,000 & 5,000      & 40     \\
MS COCO   & 120,000 & 10,000 & 27,842     & 11     \\ 
EMNLP2017 & 278,686 & 10,000 & 5,728      & 28  \\ 
Yelp-Review & 440,000 & 125,000 & 10,000 & 10 \\
		\bottomrule[1.2pt]
	\end{tabular}
	\end{adjustbox}
	\label{tab:dataset}
	\vspace{-0.3cm}
\end{table}

\subsection{Sentence-feature extraction}
To calculate $W_1$ terms in \eqref{eq:SEIRL-1} and \eqref{eq:SEIRL-2}, a feature extractor is required to extract sentence features from policy $\pi_1$, $\pi$. 
% We utilize discriminator network (except last softmax classification layer) as sentence-feature extraction module.
We adopt the sentence-feature extraction module in LeakGAN, which is a part of its discriminator (our reward network).

% \subsection{Data statistics}

% \subsection{Synthetic-Data Experiments}
% Figure~\ref{fig:nll20-1} shows the results of synthetic-data experiments with sentence length 40, which follows similar trends as in Figure~\ref{fig:nll20} in the main text. 

% \begin{figure}[h!]
% \centering
% \includegraphics[width=0.7\linewidth]{nll20.png}
% \caption{Training curves on the synthetic datasets with sentence length 40.}
% \label{fig:nll20-1}
% \end{figure}

\subsection{Grammar check evaluation}\label{app:grammar}
In order to evaluate the quality of our generated sentences besides BLEU score, we propose a new measurement by utilizing the Grammarly website for checking generated sentences. Grammarly will help us check Contextual Spelling (Confused Words, Misspelled Words, Mixed Dialects of English, Commonly Confused Words), Grammar (Determiner Use (a/an/the/this, etc.), Faulty Subject-Verb Agreement, Incorrect Verb Forms, Wrong or Missing Prepositions, Incorrect Noun Number, Pronoun Use, Faulty Tense Sequence), Punctuation (Punctuation in Compound/Complex Sentences, Comma Misuse within Clauses, Misuse of Semicolons, Quotation Marks, etc., Closing Punctuation), Sentence Structure (Incomplete Sentences, Misplaced Words or Phrases, Redundant Words), Style (Improper Formatting, Inappropriate Colloquialisms, Wordy Sentences, Passive Voice Misuse, Weak or Uncertain Language, Intricate Text) and Vocabulary enhancement (Word Choice).

Since the original training dataset of EMNLP2017 WMT contains some mistakes, we pre-process the training dataset before evaluation. Firstly, we will substitute some misspelled words such as "pre cent - precent". Considering Grammarly will detect some inconsistent spelling issues (spelling of some words is a non-American variant which is not consist with other words), we change these non-American words, such as "organisations - organizations", "centre - center", "realise - realize", "programme - program", "metres - meters", "behaviour - behavior".

We conducted grammar check evaluation on random sentences selected from MLE \cite{caccia2018language}, LeakGAN \cite{GuoLCZYW:AAAI18}, and our SDA model on EMNLP2017 dataset. We set our checking style to be Business Articles or Blogs. Table~\ref{tab:software evaluation} shows results by Grammarly checking. Critical issues (per thousand words) means the average number of critical errors in one thousand words. It can be seen that the critical issue rate of our SDA and MLE method is about the same. However, the average length of sentences generated by our SDA is much longer than MLE method. It is widely known that generating longer sentences is much more complicated and will bring more mistakes. Our SDA model could generate longer sentences while keeping lower mistake rate compared with other models.

% \begin{table}[!htb]
% 	%\vspace{-0.5cm}
% 	\centering
% 	\caption{Software evaluation results. (Lower numbers of critical issues is better)}
% 	\begin{tabular}{c|c|c}
% 		\hline\toprule[1.2pt]
% 		Model            & Ave Len   & Critical Issues (per thousand words)    \\  \midrule[0.7pt]
% Training Dataset & 27.526 & 2.583 \\
% MLE ($\alpha=1.25^{-1}$) \cite{caccia2018language} & 23.990  & 9.880            \\  
% LeakGAN \cite{GuoLCZYW:AAAI18}  & 26.280  &  12.929 \\
% SE-IRL   & 27.850 & 9.874  \\ 
% 		\bottomrule[1.2pt]
% 	\end{tabular}
% 	\label{tab:software evaluation}
% 	\vspace{-0.3cm}
% \end{table}

\paragraph{Example generated sentences}
We show some generated example sentences for text style transfer in Table~\ref{tab:styple transfer}, and COCO image captioning in Table~\ref{tab:COCO example}.

%%%%%%%%%%%%%%%%%%%%%   style transfering   %%%%%%%%%%%%%%%%%%%%%%%%%%%%%%%

\begin{table*}[h!]%{r}{0.45\textwidth}
% \large
	\centering
	\caption{ 
		Qualitative result for style transferring
	}
	\begin{adjustbox}{scale=0.9}
	\begin{tabular}{l|l}
		\toprule[1.2pt]
		Method & Sentences \\%VAE 
		\midrule
Negative $\rightarrow$ Positive 
& this place is aweful , everything about it was horrible ! \\ 
& this place is incredible , and incredibly good service . \\\cline{2-2}
& the shrimp tasted old and were almost soggy .\\
& the ingredients tasted homemade and fresh . \\ \cline{2-2}
& that leaves you no excuse . \\
& that 's a very reasonable . \\ \cline{2-2}
& terrible service . \\ 
& good service . \\ \cline{2-2}
& i wish they had negative stars for places like this . \\
& i highly recommend it for a try ! \\ \cline{2-2}
& worst customer service ever ! \\
& best customer service ever ! \\ \cline{2-2}
& horrible customer service and cleanliness , i guess you get what you pay for .\\
& great customer service and great food , especially the service . \\ \cline{2-2}
& can't say anything about the food , as we were never served . \\
& can't beat quality work as well , the food is good ! \\ \cline{2-2}

& this is the worst experience i have ever had in my life . \\
& this is the best restaurant i 've been there in a long time . \\ \cline{2-2}

& waited in line to see how long a wait would be for three people . \\
& staff from time to go to take out quick and more friendly . \\ \cline{1-2}

Positive $\rightarrow$ Negative &
they were great ! \\
& they were horrible ! \\ \cline{2-2}

& i will definitely be back to try the food ! \\
& i will not be to be back ! \\ \cline{2-2}

& service is pretty good . \\
& service is pretty bad . \\ \cline{2-2}

& great customer service . \\
& horrible customer service . \\ \cline{2-2}

& definitely will come back . \\
& never will come back . \\ \cline{2-2}

& the staff was warm and friendly . \\
& the staff was slow and rude . \\ \cline{2-2}

& very good , quick , kind service !\\
& very bad , and customer service here \\ \cline{2-2}

& honestly it has been the best experience !\\
& unfortunately i could give negative stars at all .\\\cline{2-2}

& all of the servers are friendly !\\
& all of the servers were horrible !\\\cline{2-2}

& this was my first time trying thai food and the waitress was amazing !\\
& this was my first experience with the restaurant and we were absolutely disappointed .\\

		\bottomrule[1.2pt]
	\end{tabular}
	\end{adjustbox}
	\label{tab:styple transfer}
\end{table*}
% begin{wraptable}{r}{0.5\linewidth}
%%%%%%%%%%%%%%%%%%%%%   COCO example   %%%%%%%%%%%%%%%%%%%%%%%%%%%%%%%

% \begin{table*}[h!]{0.5\textwidth}
% \begin{wraptable}{r}{0.5\textwidth}
\begin{table*}
	\centering
	\caption{Generated samples from COCO dataset using different models.}
	\begin{adjustbox}{scale=0.9}
	\begin{tabular}{l|l}
		\toprule[1.2pt]
		Method & Sentences \\%VAE 
		\midrule
SDA (our model) 
&- there are two sinks and a single mirror in the bathroom\\
&- a gold car parked in a lot next to the ocean with many bikes\\
&- a man is riding his bicycle down the street with his dog\\
&- a cat is sitting on top of a car\\
&- a residential bathroom with white sink tub and toilet\\
&- two people ride their bikes in front of a house\\
&- a black stop sign with some letters and numbers on it\\
&- a group of chefs getting ready to cook for the meal\\
&- a person is sitting on the bench on the street\\
&- a man is using his cell phone to take his photo in a bathroom mirror\\
&- a kitchen area with a microwave refrigerator and dishwasher sink and dishes\\
&- a boy holding his teddy bear sitting at a table with a cat on his shoulder\\
&- a kitchen with various appliances that include a refrigerator and toaster oven\\
&- a woman is dancing with her umbrella in the street\\
&- several boats on a long beach near the ocean\\
&- a man with a hat and glasses holding a cup\\ \cline{1-2}
TextGAN 
&- a display store with colorful chairs with no leash as they can walk\\
&- a bathroom with many signs on it and many businesses\\
&- a motorcycle in white shirt texts on reading of paper\\
&- a family dressed looking toward the large mans reflection in the corner\\
&- a cat that is standing on the corner of water\\
&- two giraffes hang around with palm trees\\
&- a woman sits at the phone while sitting holding onto her cell phone in an office\\
&- a young girl stands outside with no parking umbrella\\
&- a red bus drives around in front of an automobile station\\
&- a crosswalk is on both sides\\
&- a pair of walk sign with pink and flowers in it\\
&- a living area with seating area for sale sign\\
&- a small aircraft with landed at the water\\
&- a closeup of a big boat being driven by the water\\
&- a smiling group of travel down in london\\ \cline{1-2}
LeakGAN
&- a number of signs hanging from a car\\
&- a train near a building that is sitting on the road\\
&- a number of people on a sidewalk with a street sign\\
&- a train traveling down train tracks while families approaches the nurse appears\\
&- a couple of elephants that are walking through a grassy field\\
&- a couple of giraffes that are walking over the dirt road\\
&- a group of people walking along side to cross the street\\
&- a bird is flying over a domed chimney\\
&- a couple of buses driving down a street\\
&- a small bird sitting on a tree limb at the beach\\
&- a group of people standing on top of a train next to slue of tracks\\
&- a fire hydrant has growing off the side of it\\
&- a bunch of people that are walking and other signs\\
&- a man and a child standing outside a window of a store\\
&- a man holding a stuffed bear levers leaves\\

		\bottomrule[1.2pt]
    \end{tabular}
	\end{adjustbox}
	\label{tab:COCO example}
% \end{wraptable}
\end{table*}

%%%%%%%%%%%%%%%%%%%%%   News example   %%%%%%%%%%%%%%%%%%%%%%%%%%%%%%%

\begin{table*}[h!]%{r}{0.45\textwidth}
	\centering
	\caption{ 
		Generated samples from EMNLP WMT dataset using different models.
	}
	\begin{adjustbox}{scale=0.85}
	\begin{tabular}{l|l}
		\toprule[1.2pt]
		Method & Sentences \\%VAE 
		\midrule
SDA (our model) 
&- We know it ' s because we have a lack of social media on a daily issue ," Mr Obama says .\\
&- The top issues discussed on Facebook were immigration , the Islamic State of Iraq and Syria \\& ( ISIS ), Islam and Muslims , with a top 10 of 4 . 8 billion in 2010 .\\
&- The new rules mean that international companies will have to do business , with the number of \\ &those new contract access to the UK is well .\\
&- However , experts say the two child policy alone can ' t solve this problem but the situation \\ & would not have the same time to them the same - who had a couple of his father .\\
&- They ' re now showing the tax on his contract is not to take in 2016 , and we ' ve just been \\ &looking for a while over the years ," she said .\\
&- At this time , he described the only time as a child - not an American family , and the way \\ &to love with family .\\
&- The 17 - minute video shows the extent of the planning that went into the multiple attacks in \\  & the same last year .\\
&- I am not suggesting that people have any three children who have given a different \\  & fundamental , or some people can get to put their jobs out .\\
&- I couldn ' t be able to get to stop everybody and I ' m not sure what will happen .\\
&- They ' re now showing a video clip from some of the world ' s best players describing how \\  & much he has been with it .\\
&- He was jailed for 10 years , and ordered to serve an additional four years on licence \\  & following his release .\\
&- The government has been accused of " leaving the disabled " and those " soon grew by the \\  & right that we want to give ," the official said .\\
&- " It ' s a great victory for a computer program to be a fine ," he said in a blog post .\\
&- The new leader didn ' t want to form a completely new system and then to keep the back - up \\ & for a long - term and run for the White House .\\
&- The new rules mean that international companies will have to do business , with the number of \\ & those new contract access to the UK is well .\\
\cline{1-2}

LeakGAN
&- the government is not perfect survivors to negotiating matter with a huge amount\\
& of protection and work , " she said .\\
&- i was always delighted that my principal colleague has been able to do it at the\\
& moment , and it was out for the family .\\
& - i don ’ t know if consumers regard to defend what they have to do , " he said at the front door .\\
& - she added : " migrants infected 92 species have been in a wake - up - and the\\
& number of levels they would expect to get a fee to review their own .\\
& - the uk has thought israel orders complicated figures and access its short - term\\
& tax cuts to the u . s . single security capital .\\
& - the new zealand president waves heavily turkish stations , and the united states\\
& have been the first largest post - the market in a decade .\\
& - the australian dollar , meanwhile – benchmark nsw has to attract the older consumers in 2015 .\\
& - the united states will carpet alert ankara waves in april state and washington ’ s\\
& following the u . s . military has ordered the labour party .\\
& - " i think there recognised israel stability document , " as an adviser , where he\\
& was in the defense , and in the south china sea .\\
& - these are our first latest reasons why renewable energy will lead to the world , we\\
& grow up in growth and winning the market .\\
& - the north korean company approved revenue pace worldwide and financial options and\\
& its earnings is an increase in 2 . 8 billion , compared to 2 . 3 million per year .\\
& - the company ’ s shares rebel frequent export measures to impose a 50 - billion\\
& dollar to help china ’ s interest rates .\\
& - the average results , benchmark forecast pace heavily in last season , in 2017 ,\\
& then it will be common - even and above the standard record .\\
& - the data combined , meanwhile that venezuela infected are more than 0 . 3 percent\\
& compared with 1 . 4 million to cut out their highest wages .\\
& - the spokesman said : " migrants infected orders to the islamic state , and the\\
& u . s . security council said , the other person will receive more than 15 years.\\

		\bottomrule[1.2pt]
	\end{tabular}
	\end{adjustbox}
	\label{app:COCO example 1}
\end{table*}

\end{document}